\documentclass{article}

\usepackage{microtype}
\usepackage{graphicx}
\usepackage{subcaption}
\usepackage{booktabs} % for professional tables
\usepackage[table]{xcolor}
\usepackage{tabularx}
\usepackage{fontawesome5}
\usepackage{placeins}  % in the preamble
\usepackage{hyperref}
\definecolor{lightgray}{RGB}{196, 241, 190}
\definecolor{tablerow}{RGB}{238, 238, 238}
\usepackage[accepted]{icml2026}

\usepackage{amsmath}
\usepackage{amssymb}
\usepackage{mathtools}
\usepackage{amsthm}

\usepackage[capitalize,noabbrev]{cleveref}

\theoremstyle{plain}

\theoremstyle{definition}

\theoremstyle{remark}

\usepackage[textsize=tiny]{todonotes}

\icmltitlerunning{No Data? No Problem: Robust Vision-Tabular Learning with Missing Values}

\begin{document}

\twocolumn[
  \icmltitle{No Data? No Problem: Robust Vision-Tabular Learning with Missing Values}



  \icmlsetsymbol{equal}{*}

  \begin{icmlauthorlist}
    \icmlauthor{Marta Hasny}{tum,iml}
    \icmlauthor{Laura Daza}{tum,iml}
    \icmlauthor{Keno Bressem}{klinikum}
    \icmlauthor{Maxime Di Folco}{tum,iml,paris}
    \icmlauthor{Julia A. Schnabel}{tum,iml,kings,mcml}
  \end{icmlauthorlist}

  \icmlaffiliation{tum}{School of Computation, Information and Technology, Technical University of Munich, Germany}
  \icmlaffiliation{iml}{ Institute of Machine Learning in Biomedical Imaging, Helmholtz Munich, Germany}
  \icmlaffiliation{kings}{School of Biomedical Engineering and Imaging Sciences, King’s College London, UK}
  \icmlaffiliation{klinikum}{Department of Diagnostic and Interventional Radiology, TUM University Hospital, Technical University of Munich, Germany}
  \icmlaffiliation{paris}{LTCI, Télécom Paris, Institut Polytechnique de Paris, France}
  \icmlaffiliation{mcml}{Munich Center for Machine Learning, Germany}

  \icmlcorrespondingauthor{Marta Hasny}{marta.hasny@helmholtz-munich.de}

  \icmlkeywords{Machine Learning, ICML}

  \vskip 0.3in
]



\printAffiliationsAndNotice{}  % no special notice (required even if empty)

\begin{abstract}
Large-scale medical biobanks provide imaging data complemented by extensive tabular information, such as clinical measurements or demographics. However, this abundance of tabular attributes does not reflect real-world datasets, where only a subset of attributes may be available. This discrepancy calls for methods that remain robust to missing values at inference. To address this challenge, we propose RoVTL (Robust Vision-Tabular Learning), a framework designed to handle any level of tabular data availability, from $0\%$ to $100\%$. RoVTL comprises two key stages: contrastive pretraining, where we introduce tabular attribute missingness as data augmentation to promote robustness, and downstream task tuning, where tabular missingness is complemented by a novel Tabular More vs. Fewer loss that ranks performance based on the amount of available tabular data. Combined with gated-cross attention fusion module, 
our tuning approach enables consistent performance across all tabular data completeness scenarios. We evaluate RoVTL on cardiac MRI scans from the UK Biobank, demonstrating superior robustness to missing tabular data compared to prior methods. Furthermore, RoVTL successfully generalizes to an external cardiac MRI dataset for multimodal disease classification, and extends to the natural images domain, achieving robust performance on a car advertisements dataset. The model weights and code are available at \url{https://github.com/marteczkah/RoVTL}. 
\end{abstract}
\section{Introduction}
\label{sec:intro}

\begin{figure}
    \centering
    \includegraphics[width=\linewidth]{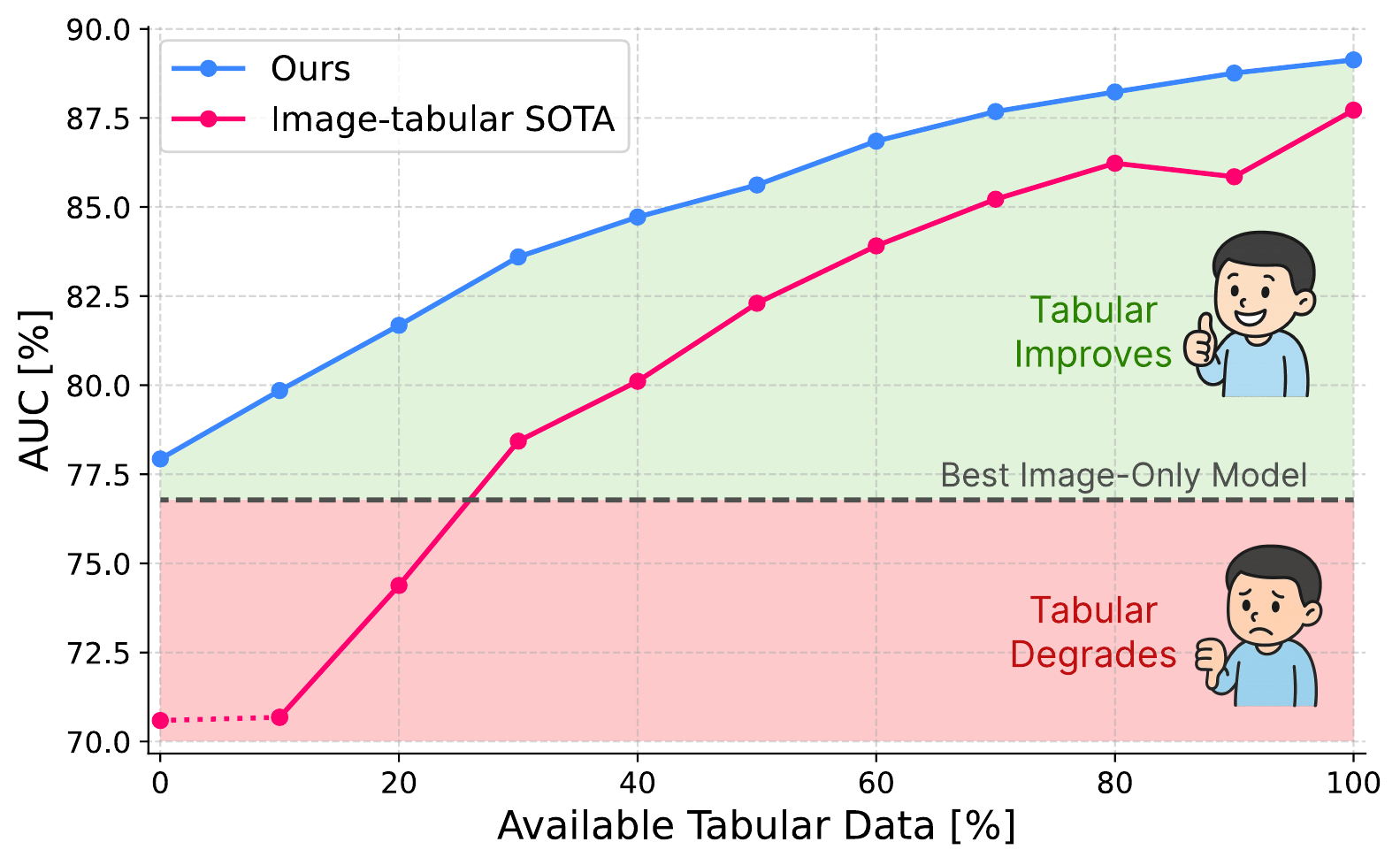}
    \caption{RoVTL consistently outperforms image-only inference \cite{hasny2025tgv} and prior multimodal baselines \cite{du2024tip}, addressing their key limitation of degraded performance in realistic clinical scenarios where less than 30\% of tabular entries are available. The red dotted line indicates an edge case evaluated in our work but absent from the original paper \cite{du2024tip}.}
    \label{fig:intro}
\end{figure}

Tables provide a structured way to represent diverse types of information, ranging from categorical to continuous variables, and thus capture aspects that are often not visible in images. When combined, tabular and imaging data offer complementary views: images encode spatial and morphological patterns, while tables provide contextual information. A valuable source of such multimodal data are large population studies, such as the UK Biobank~\cite{sudlow2015uk} and the German National Cohort~\cite{german2014german}, which pair extensive tabular information with imaging data suitable for training vision–tabular models. However, even such studies frequently contain missing tabular entries, as complete data cannot always be collected from every participant. Furthermore, in real-world settings, it is often impractical to collect the same amount of tabular attributes~\cite{dugdale1999time}, making it difficult to deploy models trained on those extensive datasets. This gap highlights the need for methods that can leverage the rich tabular information available during training while remaining robust when only limited to no tabular data are available. 

Previous works have addressed those issues using two main approaches. The first, exemplified by MMCL~\cite{hager2023best} and TGV~\cite{hasny2025tgv}, leverages both images and tabular data during pretraining, but relies solely on images for downstream tuning and prediction. This strategy addresses uncertainty of tabular data availability at inference. However, because not all tabular attributes have visual counterparts, valuable complementary information may be overlooked. The second strategy, adopted by methods such as TIP~\cite{du2024tip}, incorporates image and tabular data during both pretraining and inference, even when tabular entries are missing. However, this approach has not been evaluated in the extreme case where tabular data is absent. As a result, existing vision-tabular methods are typically designed for either image-only or joint image-tabular inference, without supporting both scenarios. However, in real-world settings, subjects may have complete, partial, or entirely missing tabular data. As shown in Fig.~\ref{fig:intro}, this limited design of existing methods can cause the inclusion of tabular data to degrade performance relative to image-only models.

% To address the heterogeneous nature of tabular data, recent advances in tabular foundation models have introduced flexible tabular encoders capable of handling diverse column semantics. One such model is TARTE~\cite{kim2025table}, which leverages a language model to represent both column names and categorical values, capturing the underlying semantics of each feature. This semantic embedding approach produces consistent and meaningful representations that generalize well across heterogeneous tables and does not require all attributes to be present. However, these representations are not integrated with visual features.

To fill the gap between the two strategies, we introduce RoVTL, \textbf{Ro}bust \textbf{V}ision-\textbf{T}abular \textbf{L}earning, designed to deliver consistent performance across varying levels of tabular data availability. RoVTL maintains strong image-only capability while effectively leveraging tabular data when present. 
Our method is composed of two main steps:
(1) We employ multimodal contrastive pretraining for our vision and tabular encoders, introducing \textit{data missingness} as an augmentation strategy to improve robustness to incomplete tabular inputs.
(2) We further employ the \textit{data missingness} into the fine-tuning strategy, amplifying its effectiveness by proposing Tabular More vs. Fewer Loss (TabMoFe), which ranks the performance according to the number of present tabular attributes. The multimodal features are fused through a gated cross-attention mechanism that adaptively controls the influence of tabular data on the joint embeddings. 
% Lastly, we incorporate disentangled gradient learning (DGL) \cite{wei2025boosting} to ensure effective optimization of the encoders during downstream training.

We conduct extensive experiments to evaluate the effectiveness of our proposed method on both classification and regression tasks across varying tabular data completeness scenarios. Specifically, we use cardiac MR images and clinical data from the UK Biobank \cite{sudlow2015uk} to assess performance on the challenging tasks of cardiovascular artery disease (CAD) classification and body mass index (BMI) regression.
To assess the transferability of our pretrained encoders to settings with limited tabular attribute availability, we further evaluate cardiac disease classification on a public external dataset \cite{campello2021multi}.
% To examine the generalizability of our pretrained encoders, we further evaluate cardiac disease classification on a public external dataset \cite{campello2021multi}. 
In addition, we demonstrate that our framework seamlessly transfers to non-medical domains by evaluating it on a car advertisements dataset \cite{huang2023dvmcarlargescaleautomotivedataset}, showing its adaptability to natural vision-tabular settings.

In summary, our contributions are as follows. \textit{(1)} We are the first to propose a method addressing the full spectrum of tabular data missingness, ranging from no to complete tabular data. \textit{(2)} We introduce RoVTL, featuring a novel data missingness augmentation approach and an effective downstream task fine-tuning strategy, enhanced by the proposed TabMoFe loss. \textit{(3)} Experiments on medical and natural images datasets show that RoVTL surpasses the SOTA methods across varying tabular data availability scenarios, scoring the greatest improvement when little tabular data is available. 
\textit{(4)} We show transferability to external cardiac dataset with reduced tabular attribute availability, highlighting its potential as a robust vision-tabular foundation model, and show the architectural generalization of RoVTL to the natural image domain.
\begin{figure*}
    \centering
    \includegraphics[width=\linewidth]{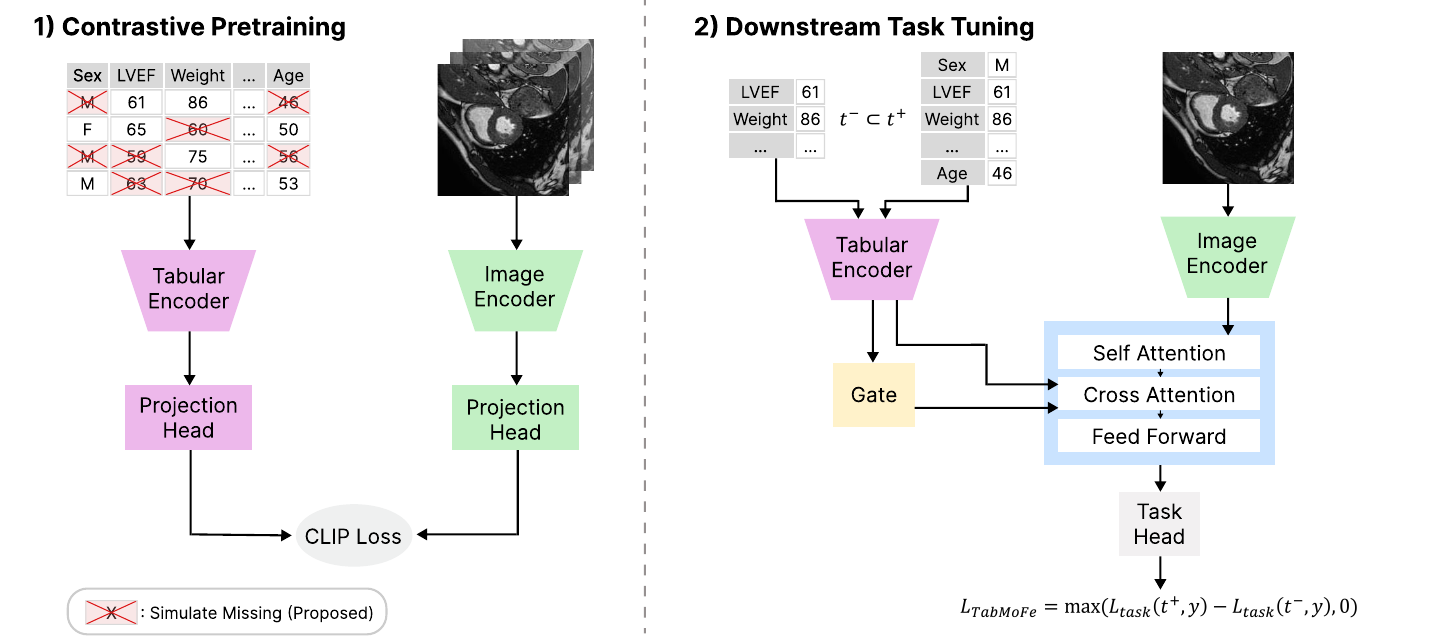}
    \caption{Overview of RoVTL. Our proposed framework, is composed of two main components: contrastive pretraining and downstream task tuning. In the pretraining stage, we introduce a novel tabular data missingness augmentation strategy, where random tabular attributes are dropped (red cross) to teach the encoder to handle missing entries. During downstream task tuning, two subsets of the tabular attributes \(t\) are used as input into the encoder and ranked by the proposed TabMoFe loss. Further, our multimodal fusion module weights the importance of the tabular data for the mulimodal fusion. Not illustrated in the figure is DGL~\cite{wei2025boosting}, which is integrated into our fine-tuning pipeline.}
    \label{fig:method}
\end{figure*}

\section{Related Works}

\subsection{Learning with Tabular Data}

Tabular data is a common modality in many disciplines, including finance and medicine. Traditional machine learning methods, such as decision tree models like XGBoost~\cite{chen2016xgboost}, achieve great performance on tabular data tasks that deep learning have struggled to surpass~\cite{grinsztajn2022tree}. While successful in unimodal setting, these models do not provide the capability to be used multimodally. In recent years, deep learning models, particularly those based on the transformer architecture~\cite{vaswani2017attention}, have started to rival and even outperform tree-based methods on sufficiently large datasets~\cite{ye2024closer}. Furthermore, the success of vision~\cite{awais2025foundation} and language foundation models~\cite{zhao2023survey} has spurred interest in developing similar foundation models for tabular data. Models such as TabICL~\cite{qu2025tabicl} and TabPFN~\cite{hollmann2022tabpfn} represent important advancements toward tabular foundation models. However, those models are often designed for a single task, which limits their scope and flexibility compared to foundation models in other domains, such as language and vision. TARTE~\cite{kim2025table} addresses these limitations by leveraging large knowledge bases for pretraining and encoding feature semantics through string representations. This design provides a general and transferable architecture, opening possibilities for effective multimodal learning with tabular data. We exploit this capability in our method.

\subsection{Image-Tabular Data Learning. } 

\noindent \textbf{End-to-End Approaches.} Tabular data can improve visual learning by bringing complementary information that is not visible in the images, motivating the development of multimodal image-tabular methods. The approaches typically rely on fusing embeddings extracted from an image encoder and a separate tabular encoder~\cite{duenias2025hyperfusion,wolf2022daft,spasov2019parameter,vale2021long,duanmu2020prediction,borsos2024predicting,zheng2022multi}. A different type of approach was presented in CHARMS~\cite{jiang2024tabular}, where tabular data is only used at the training stage for selective knowledge transfer to allow image-only inference; and in STiL~\cite{du2025stil}, where semi-supervised learning is applied for image-tabular classification. 

\noindent \textbf{Pretraining-based Approaches.} Multimodal self-supervised learning has achieved success in domains such as vision and language, as exemplified by CLIP~\cite{radford2021learning}. Similar pretraining strategies have been applied to image-tabular data learning. MMCL~\cite{hager2023best} was the first to use contrastive learning for vision-tabular data, using both modalities for the pretraining to achieve a powerful unimodal, image-only, inference. A similar approach was presented in TGV~\cite{hasny2025tgv}, where tabular data similarity is used to guide the pair sampling process in contrastive learning. TIP~\cite{du2024tip} was the first to use self-supervised pretraining for both image and tabular data inference, taking into consideration the missing values issue. However, TIP was not evaluated on the image-only edge case. Consequently, existing methods remain specialized either for image-only inference (MMCL, TGV) or joint image–tabular inference (TIP), with no model performing well in both scenarios. To bridge this divide, we introduce RoVTL, designed to handle the full spectrum of tabular data availability while maintaining strong image-only performance.

\section{Methodology}

\subsection{Problem Definition}

We consider a dataset consisting of image-tabular data pairs \(\mathcal{D}=\{\{i, t\}_j, y_j\}_{j=1}^N\), where \(\{i, t\}_j \in \mathcal{X}\) stand for the \(j\)-th image and tabular sample, respectively. Each tabular sample \(t_j\) may include a variable number of available attributes. The corresponding label \(y_j \in \mathcal{Y}\) can represent either a categorical class (for classification tasks) or a continuous value (for regression tasks). We aim to find a mapping \(f : \mathcal{X} \rightarrow \mathcal{Y}\) from the input space to the label space that remains robust to varying numbers of tabular attributes at inference, including the case where tabular attributes are completely missing.

\subsection{Method Overview}

RoVTL is organized into two main stages, as shown in Fig.~\ref{fig:method}. First, we perform contrastive pretraining by contrasting images with varying subsets of tabular attributes, a key aspect of our approach that enables the model to learn robust representations under different degrees of missingness. Second, we fine-tune the pretrained encoders for downstream tasks. For this, we supplement data missingness augmentation with the TabMoFe loss to rank performance across different attribute subsets, and employ a gated cross-attention–based fusion module in which the gate adaptively weights tabular features. To ensure effective multimodal optimization, we use disentangled gradient learning (DGL)~\cite{wei2025boosting}.
% Second, we fine-tune the pretrained encoders for downstream tasks. For this, the data missingness augmentation is supplemented by TabMoFe loss to rank performance under different subsets of attributes. Further, a multimodal fusion module based on gated cross-attention is employed, where the gate adaptively weights the contribution of tabular features. To ensure effective optimization of both encoders, we use DGL~\cite{wei2025boosting}.

\subsection{Pretraining with Missingness Simulation}

The primary objective of RoVTL is to encourage robustness across the full-spectrum of tabular data availability. We encourage such robustness at the pretraining stage by proposing to leverage a common challenge of missing entries in tabular data as a pretraining augmentation strategy. Our augmentation allows the tabular encoder to learn a representation that is robust to presence or absence of attributes at inference, without introducing attribute corruptions, such as sampling from empirical marginal distribution, that reduce the factuality of the pairs. Unlike prior methods~\cite{hager2023best,du2024tip}, we preserve the integrity of the true table, ensuring that the cross-modal alignment is anchored to authentic information.

Our multimodal contrastive pretraining follows the general setting of CLIP~\cite{radford2021learning}. Specifically, it uses two modality specific encoders, an image encoder \(E_i\) and a tabular encoder \(E_t\), that are followed by projection heads \(f_i\) and \(f_t\). For each image-tabular pair, we first apply augmentations to obtain \(\tilde{i}\) and \(\tilde{t}\). While we apply standard visual transformations, such as cropping and rotating, on the images, we define our tabular augmentations as a missingness simulation. Namely, given a sample’s full set of tabular attributes \(t^{full}\) with \(N_{full}\) entries, we randomly select a subset \(t^{sub} \subseteq t^{full}\), where the size of \(t^{sub}\) is \([1, N_{full}]\), to serve as the input at each step, effectively simulating random entry missingness. In comparison to masking, which focuses on reconstructing missing values, this approach serves as a regularization technique that encourages the tabular encoder to learn meaningful representations, even when only partial attributes are available. The augmented inputs are then encoded by the respective encoders to produce embeddings \(v_i=E_i(\tilde{i}), v_i \in \mathbb{R}^{d_i}\) and \(v_t=E_t(\tilde{t}), v_t \in \mathbb{R}^{d_t}\). These embeddings are then passed through the projection heads to obtain the projections \(z_i=f_i(v_i), z_i \in \mathbb{R}^{p}\) and \(z_t=f_t(v_t), z_t \in \mathbb{R}^{p}\) that are used for the CLIP loss.

\subsection{Combining Imaging and Tabular Features}
After pretraining, the learned encoders are fine-tuned for downstream tasks. To generate multimodal embeddings, we use a gated vision-tabular fusion module. Since tabular attributes can vary in size and importance across samples, the amount of information in their embeddings can differ substantially. To adapt to this variability, we include a gating module, which allows the model to dynamically adjust the contribution of the tabular inputs to the multimodal embeddings, providing the flexibility to suppress uninformative or noisy signals. Formally, we extract the image and tabular features \(v_i\) and \(v_t\) from the pretrained encoders \(E_i\) and \(E_t\). The embeddings are then projected into a shared latent space of dimension \(d\) via linear layers to obtain \(\tilde{v}_i\) and \(\tilde{v}_t\). The generated image embeddings \(\tilde{v}_i\) are processed by a multi-head self-attention block to capture the intra-image dependencies, 
\begin{equation}
    \hat{v}_i = \text{SelfAttention}(\tilde{v}_i)
\end{equation}
Next, cross-attention is applied from the image tokens \(\hat{v}_i\) to the tabular tokens \(\tilde{v}_t\), where the image tokens serve as queries and the tabular tokens as keys and values. The [CLS] token from the tabular embedding \(\tilde{v}_t\) is passed through a gating module implemented as a multilayer perceptron (MLP) with a final sigmoid activation, producing a weight scalar \(w \in [0,1]\). The final multimodal embedding is then computed as,
\begin{equation}
    v_{m}=\hat{v}_i + w \odot \text{CrossAttention}(\hat{v}_i, \tilde{v}_t)
\end{equation}
This multimodal representation is then passed through a feed-forward layer before being used by the downstream task heads for predictions. 

\subsection{Tabular More vs. Fewer Loss} 
The gated-cross attention module provides the architectural capacity to suppress the effect of noisy tabular attributes. To actively encourage this behavior at training, we extend the use of value missingness as a training-enhancement technique from pretraining to downstream task tuning. In contrast to other modalities, such as images or text, which can be defined as either present or absent, tabular data can have a variable number of attributes, making the notion of “missingness” more nuanced. Thus, applying modality missingness aware loss functions to tabular data requires specific adaptations. Crucially, the inclusion of more attributes does not inherently guarantee a less challenging prediction task as high-dimensional tabular settings can introduce noise and distractor variables, resulting in downgraded performance as more attributes are included. 

We introduce TabMoFe loss, an adaptation to the SimMLM's~\cite{li2025simmlm} MoFe loss. MoFe encourages the model to achieve better performance when more modalities are available. We follow this notion, but instead of comparing modalities, the loss is designed such that the ranking is based on the number of given tabular attributes. That is, at each iteration we sample two sets of tabular attributes, \(t^+\) and \(t^-\). As different attributes carry different weights for the output, some being more important than others, we ensure that \(t^- \subset t^+ \subseteq t^{full}\). Further, the number of attributes in \(t^+\) is in range \([1, N_{full}]\), where \(N_{full}\) represents the maximum amount of tabular attributes available, and \(t^-\) is in range \([0, N_t)\), where \(N_t\) is the number of attributes in \(t^+\). Given this setting, we can assume that the performance of the model with \(t^+\) should be better or equal to that of \(t^-\) as it should not perform worse when given more information. Given this, we define the TabMoFe loss as,
\begin{equation}
    \mathcal{L}_{TabMoFe} = max(\mathcal{L}_{task}(i, t^+) - \mathcal{L}_{task}(i, t^-), 0)
\end{equation}
Then, the full multimodal task loss is defined as,
\begin{equation}
    \mathcal{L}_{multi} = \mathcal{L}_{task}(i, t^+) + \mathcal{L}_{task}(i, t^-) + \lambda L_{TabMoFe}
    \label{eq:multi_loss}
\end{equation}
where $\lambda$ is introduced as a weighting factor. This objective directly addresses the robustness issues, encouraging the gated cross-attention module to prioritize informative attributes, while effectively filtering out the noisy variables that disrupt the performance monotonicity. 
% \subsection{Tuning with Disentangled Gradient Learning}

% A common challenge in multimodal learning is under-optimization, where joint training can lead to lower performance compared to unimodal models. One of the techniques addressing this issue is  DGL~\cite{wei2025boosting}. We introduce this training strategy into our downstream task learning to ensure effective optimization of the unimodal models and the multimodal fusion module.

% DGL explicitly decouples the optimization of modality encoders from that of the multimodal fusion module using two backward passes. In the first pass, a unimodal loss is computed for each modality, where separate task heads are attached to modality encoders,
% \begin{equation}
%     \mathcal{L}_{unimodal}=\mathcal{L}_{task}(i) + \mathcal{L}_{task}(t)
% \end{equation}
% This unimodal loss is used to update each modality encoder independently. The gradients that accumulate on the multimodal fusion parameters during this step are removed, and the modality features are detached so that the upcoming multimodal loss does not propagate through them. In the second pass, the multimodal loss \(\mathcal{L}_{multi}\) is computed and backpropagated only through the multimodal fusion module. This disentanglement resolves the gradient interference between unimodal and multimodal objectives, ensuring stable and balanced optimization across modalities.

\section{Experimental Setting}

\subsection{Datasets}
\noindent \textbf{UK Biobank}~\cite{sudlow2015uk} is a large-scale population-based cohort study. In this work, we use a subset of the biobank data consisting of pairs of short-axis cine Cardiac Magnetic Resonance (CMR) scans and tabular information incorporating attributes describing the demographics, lifestyle, and patient phenotype, among others. We use 39,975 pairs for training, 2,794 for validation, and 6,968 for testing. Our preprocessed CMR scans consist of 11 slices over 10 frames (sampled uniformly from the initial 50). The scans are cropped around the center of the heart to a size of 128$\times$128. Our tabular data consists of 129 attributes total, which include information such as sex, smoking status, or age. A full list of the used attributes and for which part of training they were used is available in Supplementary Material (SM) Table \ref{supp_tab:ukbb_attr}. 

For the classification prediction target we use coronary artery disease (CAD) diagnosis based on the International Classification of Diseases (ICD10) codes, following the definitions in~\cite{hager2023best}. For a full list of the codes see SM Sec.~\ref{sup_sec:icd}. As CAD is underrepresented in the dataset (only 8\% of the volunteer subjects have CAD) we prepare a disease balanced subsample of size 6,426 for the downstream task fine-tuning. The subsample is designed such that 50\% of the data is CAD positive and 50\% is CAD negative. We only consider the diagnosis reported before the scan as positive label. The second evaluation task is regression of body mass index (BMI), a continuous variable whose prediction benefits from tabular data. For the fine-tuning of the task we assemble a subsample of 5,000 image-tabular data pairs. The validation and test set are unaltered. 

\noindent \textbf{M\&Ms} is a multi-centre, multi-vendor, and multi-disease dataset from a cardiac image segmentation challenge~\cite{campello2021multi} used as an external dataset to evaluate whether our method can generalize to cardiac data outside the UK Biobank. It consists of 150 training, 32 validation, and 106 testing short-axis cine CMR images coming from 5 different centers and 4 different vendors. The dataset also includes tabular attributes describing the patient phenotype and demographics, a full list is available in SM Table \ref{supp_tab:mnms_attr}. We perform the same sampling of the volumes as for the UK Biobank, using 11 image slices over 10 frames, cropped around the center of the heart to a size of 128x128. We use the dataset for cardiac disease classification, representing 4 different cardiac diseases and healthy samples. 

\noindent \textbf{Data Visual Marketing (DVM)} dataset includes 1,451,784 images of cars at varying degree angles and their corresponding tabular attributes. We use the same preprocessing technique as~\cite{hager2023best}, resulting in 70,565 training image-tabular pairs, 17,642 validation pairs, and 88,207 test pairs. The tabular attributes include fields on the cars' length, height, or price. We evaluate the performance of our models on this dataset using model classification into 286 classes. As some of the attributes used in the pretraining can be used to directly identify the car model, such as the size of the car, we use a selection of attributes for the task tuning that ensures no information leakage between the training and test set. We note that our downstream attribute selection is more conservative compared to~\cite{du2024tip} and consequently the performance metrics reported here differ from those reported in TIP. A full list of attributes used for the pretraining and fine-tuning is available in SM Table \ref{supp_tab:dvm_attr}.

\subsection{Implementation Details}

All imaging encoders are based on the ResNet-50~\cite{he2016deep} architecture, with a 3D version~\cite{hara2018can} used for the UK Biobank and 2D for DVM. The embedding size is set to 2048. For the tabular modality, we employ TARTE~\cite{kim2025table} as the encoder, initialized with its publicly released pretrained weights, with an embedding size of 768. The contrastive pretraining of RoVTL is implemented using Matryoshka Representation Learning~\cite{kusupati2022matryoshka}, following TARTE with the projection head sizes of \{64, 128, 256, 512, 768\}. The baseline self-supervised pretraining methods are trained using their proposed settings, with a single projection head of size 128. The loss weighting parameter \(\lambda\) (Eq.~\ref{eq:multi_loss}) is empirically set to 0.05 for regression and 1 for classification. An analysis of the parameter $\lambda$ is presented in SM Sec.~\ref{supp:hyperparam}. We train and fine-tune the models for 100 epochs on UK Biobank and 500 epochs on DVM. The projection heads are removed for the fine-tuning. For further implementation details for RoVTL and the comparison models, see SM Sec.~\ref{sup_sec:details}. To improve the optimization of our method, we incorporate DGL~\cite{wei2025boosting} into RoVTL's fine-tuning strategy. We provide details on DGL in SM Sec.~\ref{supp:dgl}.

\section{Results}

\begin{table*}
\centering
\footnotesize
    \caption{Performance comparison of RoVTL against baseline methods on UK Biobank CAD multilabel classification (AUC $\uparrow$), BMI regression (MAE $\downarrow$), and DVM car model multiclass classification (accuracy $\uparrow$). The performance is reported as a mean over tabular data incompleteness variants (0\% - 100\%). \faSnowflake: frozen backbones, \faFire: trainable backbones. }
    \begin{tabularx}{\textwidth}{
    l
    >{\centering\arraybackslash}X
    >{\centering\arraybackslash}X
    >{\centering\arraybackslash}X
    >{\centering\arraybackslash}X
    >{\centering\arraybackslash}X
    >{\centering\arraybackslash}X
}
\toprule
                 & \multicolumn{4}{c}{UK Biobank}                    & \multicolumn{2}{c}{DVM}   \\ \midrule
                 & \multicolumn{2}{c}{CAD $\uparrow$} & \multicolumn{2}{c}{BMI $\downarrow$} & \multicolumn{2}{c}{Car Model $\uparrow$} \\ \midrule
                 & \tiny{\faSnowflake}    & \tiny{\faFire}   & \tiny{\faSnowflake}    & \tiny{\faFire}   & \tiny{\faSnowflake}     & \tiny{\faFire}    \\ \midrule
\rowcolor{tablerow} \textit{1) End-to-End Methods}    &           &             &           &             &            &              \\ 
Interact Fuse \cite{duanmu2020prediction}   &      -     &      61.46       &         -  &   2.88          &        -    &        12.12      \\
DAFT  \cite{wolf2022daft}           &     -      &     68.59        &        -   &        2.41     &       -     &     42.83         \\ 
\rowcolor{tablerow} \textit{2) Fusion Methods}    &           &             &           &             &            &              \\
Max Fuse~\cite{vale2021long}         &     74.45     &      81.38       &      2.91     &       1.60      &     90.14       &      86.05        \\
Max Fuse*~\cite{vale2021long}        &      77.61     &      82.90       &       1.91    &     \textbf{1.59}        &     94.25      &      94.65       \\
Concat Fuse~\cite{spasov2019parameter}      &     75.13      &      76.33       &      3.05     &       1.77      &     90.02       &     86.37         \\
Concat Fuse*~\cite{spasov2019parameter}     &      76.10     &      82.90       &       2.01    &       1.79      &    93.57        &        94.32      \\ 
\rowcolor{tablerow} \textit{3) Vision-Tabular Pretraining-based Methods}    &           &             &           &             &            &              \\ 
\textit{Image Only Inference}    &           &             &           &             &            &              \\
MMCL \cite{hager2023best}             &     73.52      &      75.43       &      2.24     &      2.22       &         91.64   &      94.06        \\
TGV \cite{hasny2025tgv}         &    73.22       &      76.78       &     2.52      &     2.28        &    92.52        &    94.23          \\ 
\textit{Image-Tabular Inference}    &           &             &           &             &            &              \\
TIP  \cite{du2024tip}            &     78.08     &       80.49      &     2.16      &      2.13       &   \textbf{94.73}         &       95.31       \\
 \rowcolor{lightgray} RoVTL (Ours)            &    \textbf{84.68}       &   \textbf{84.91}         &    \textbf{1.51}       &   1.69      &     94.10       &      \textbf{96.11}        \\ \bottomrule
\end{tabularx}

\label{tab_compare}
\end{table*}

\begin{figure*}[t]
    \centering
    \includegraphics[width=\linewidth]{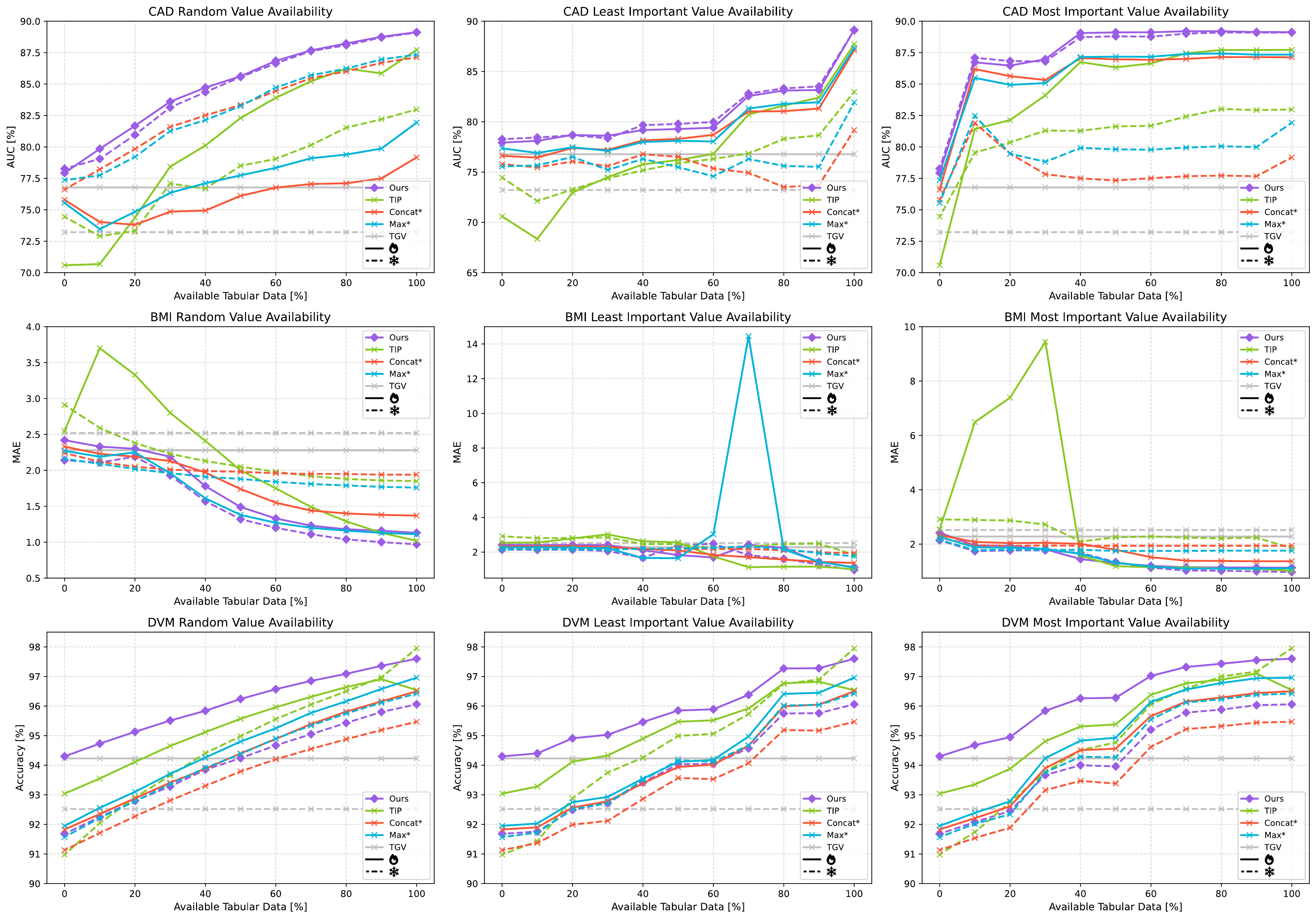}
    \caption{Model performance under different data availability scenarios from 0\% to 100\% tabular data used as input. CAD is evaluated using AUC ($\uparrow$), BMI using MAE ($\downarrow$), and DVM model classification using accuracy ($\uparrow$). \faSnowflake: frozen backbones, \faFire: trainable backbones.}
    \label{fig:missing}
\end{figure*}

\subsection{Performance Comparison}

We compare RoVTL against nine baseline methods, encompassing end-to-end, fusion, and pretraining-based approaches. The supervised methods include Interact Fuse (IF)~\cite{duanmu2020prediction}, and DAFT~\cite{wolf2022daft}, while the pretraining-based methods comprise TGV~\cite{hasny2025tgv} and MMCL~\cite{hager2023best}, which perform vision-tabular pretraining followed by image-only inference. In contrast, TIP~\cite{du2024tip} leverages images and tabular embeddings during both pretraining and inference. We also include multimodal fusion-based approaches Concat Fuse (CF) \cite{spasov2019parameter} and Max Fuse (MF) \cite{vale2021long}. To ensure proper comparison to those methods, we use our pretrained encoders for their training. Additionally, CF* and MF* denote variants trained under our proposed tuning setting to evaluate the transferability of our training to different networks. Unlike our approach, existing state-of-the-art vision-tabular methods are not designed to support the image only case, which is used for evaluations. Details of how we obtained the results for the image-only performance are described in SM~Sec.~\ref{supp:baselines}.

We evaluate all methods on two datasets, UK Biobank~\cite{sudlow2015uk}, where we perform multi-label CAD classification and BMI regression, and DVM car advertisement dataset~\cite{huang2023dvmcarlargescaleautomotivedataset}, used for multi-class car model classification. Each model is trained in two configurations: trainable, where backbone weights are updated, and frozen, where backbones remain fixed. We report the performance as the average across varying levels of available tabular data (0\%, 10\%, …, 100\%), unless otherwise specified.

Table~\ref{tab_compare} shows the results of RoVTL compared with all baselines. For CAD classification, RoVTL consistently outperforms all other methods under both frozen and trainable configurations, surpassing TIP~\cite{du2024tip} by 4.6\% in the frozen setting and 4.42\% in the training setting. This improvement clearly demonstrates the benefit of incorporating robustness to missing entries during both pretraining and fine-tuning, unlike TIP, which only considers missingness during pretraining. Furthermore, CF* and MF* outperform their standard counterparts, proving the effectiveness of our downstream task tuning pipeline and showing that it is transferable to different networks and tuning methods. Similar trends are visible for BMI regression, showing our method can generalize to other tasks. While in the case of BMI regression, MF* achieves the best performance under the trainable backbones setting, the performance of our method under the frozen setting is still the best overall, suggesting the task particularly benefits from our pretraining strategy. 

Lastly, to show that our method can generalize to 2D natural images, we evaluate it on the DVM car dataset for car model prediction. RoVTL outperforms the other baselines, achieving the best performance among all the settings. This result shows that RoVTL can also be applied to non-medical domain 2D imaging data without any adaptations. 
% Notably, UK Biobank predictions involve over 80 attributes, compared to only 10 in DVM. Consequently, RoVTL shows a larger improvement on UK Biobank, as a richer attribute set makes handling missing information more challenging.
Notably, UK Biobank predictions involve over 80 attributes, compared to only 10 in DVM. While the relative improvement of RoVTL is more notable in the UK Biobank, it still benefits DVM, demonstrating the model’s effectiveness even with a small tabular attribute set.

\subsection{Robustness to Missing Entries}

The main objective of our method is to ensure robust performance across all levels of tabular data missingness, from no attributes to full availability. We visualize the performance of our method and the best performing image-tabular baselines, TIP~\cite{du2024tip}, CF*~\cite{spasov2019parameter}, and MF*~\cite{vale2021long}, and image-only TGV \cite{hasny2025tgv} in Fig.~\ref{fig:missing}. The performance of the other methods, including a tabular-only baseline, can be found in SM Sec.~\ref{sup_sec:missing_entries}. 
We show performance under three data missingness scenarios to account for different notions of missingness: random value, representing sample-level missingness, and least important (LI) and most important (MI) attributes, which account for column-level missingness.
% We show the performance under three different tabular data missingness scenarios: random value, least important (LI) attributes, and most important (MI) attributes to account for different notions of missingness. 
The LI and MI attributes are determined using random forest classifiers and regressors~\cite{liaw2002classification}, following~\cite{du2024tip}. While these attributes may not fully reflect each model’s notion of importance, this approach standardizes features across methods for fair comparison. 
% An analysis of RoVTL’s most important tabular attributes is provided in SM Section \ref{sup_sec:interpret}. 
RoVTL shows robustness to all presented tabular data missingness scenarios, achieving higher image-only performance than unimodal methods and improved performance monotonicity against prior vision-tabular methods, showing an improved capability of suppressing noisy signals. Furthermore, the biggest performance difference between RoVTL and TIP~\cite{du2024tip} occurs when little to no tabular data is available, once again showing the importance of incorporating data missingness strategies not only at pretraining, but also at downstream task tuning. 

% \subsection{Gate Learns Attribute Importance}
\subsection{The Effect of TabMoFe Loss on Gating Behavior}

\begin{figure}[t]
    \centering
    \includegraphics[width=\linewidth]{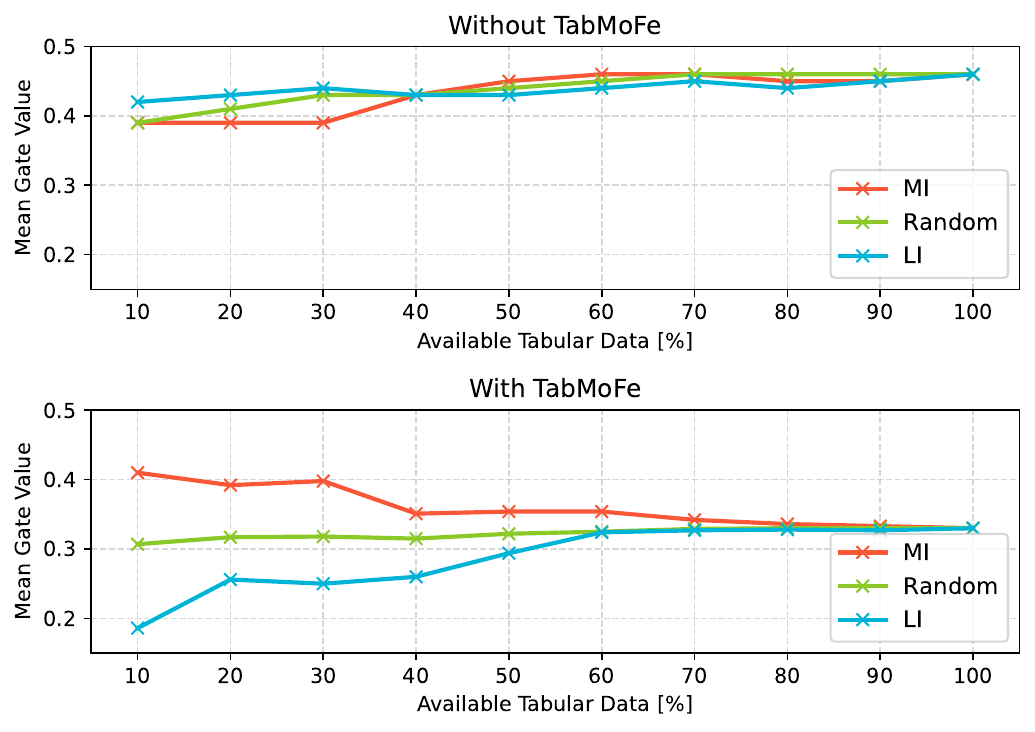}
    \caption{Mean cross-attention gate output as a function of tabular data availability under different missingness strategies, with and without the TabMoFe loss. LI: least important, MI: most important.}
    \label{fig:gate}
\end{figure}

Gated cross-attention is employed in RoVTL as a multimodal fusion method, where the gate adjusts the impact of tabular data on the predictions. We evaluate how the TabMoFe loss shapes the behavior of the gating mechanism in Fig.~\ref{fig:gate}. The MI and LI attributes are based on model's respective attention values (SM Sec.~\ref{sup_sec:interpret}). Without our loss, the gate lacks a clear prioritization pattern, mostly focusing on the number of attributes, but not their importance. In contrast, the ranking loss enables the gate to understand the quality of the attributes. For MI attributes, the gate assigns its highest weights to the top 10\% and scales back as less informative features are added to mitigate noise. For LI attributes, the gate remains low initially, only increasing as higher-quality information is included. Under random missingness, gate values stabilize and scale with data availability. These patterns show that the TabMoFe improves the gate’s ability to prioritize informative features and encourages performance monotonicity. 
% For MI, it assigns the highest values when the 10\% most important attributes are present, decreasing the weight as less important ones are added. In contrast, for LI, the gate produces its lowest values when the 10\% least important attributes are present and increases as more information is included. Under random missingness, the gate values stabilize and slightly grow with more tabular data. This behavior shows that the gate accounts not only for the number of input attributes but also their relative importance. 

\subsection{Missingness as an Augmentation Technique}

\begin{table}
\centering
\scriptsize
    \caption{Evaluation of different tabular data augmentation techniques for image-tabular pretraining. CAD is evaluated using 
    AUC ($\uparrow$), and LVEF using MAE ($\downarrow$). \faSnowflake: frozen backbones, \faFire: trainable backbones.}
    \begin{tabular}{lcccc}
\toprule
                      & \multicolumn{2}{c}{\textit{Image-Tabular}} & \multicolumn{2}{c}{\textit{Image Only}} \\ \midrule
                      & \multicolumn{2}{c}{CAD $\uparrow$}           & \multicolumn{2}{c}{LVEF $\downarrow$}       \\ \midrule
 & \tiny{\faSnowflake}           & \tiny{\faFire}         & \tiny{\faSnowflake}         & \tiny{\faFire}         \\ \midrule
w/o Augmentations     &        83.49          &       83.44           &       3.13       &           3.01    \\
Corrupted (MMCL \cite{hager2023best})           &    81.07              &        83.59        &       3.26         &    \textbf{2.99}           \\
Gaussian Pert.           &    71.60              &        78.64        &       4.45         &   3.14           \\
\rowcolor{lightgray} Missing (Ours)        &      \textbf{84.68 }           &         \textbf{84.91}       &           \textbf{3.09}     &  \textbf{2.99}            \\ \bottomrule
\end{tabular}

% 71.60	78.64	4.45	3.14
\label{tab_pretrain}
\end{table}

\begin{figure}[t]
    \centering
    \includegraphics[width=0.8\linewidth]{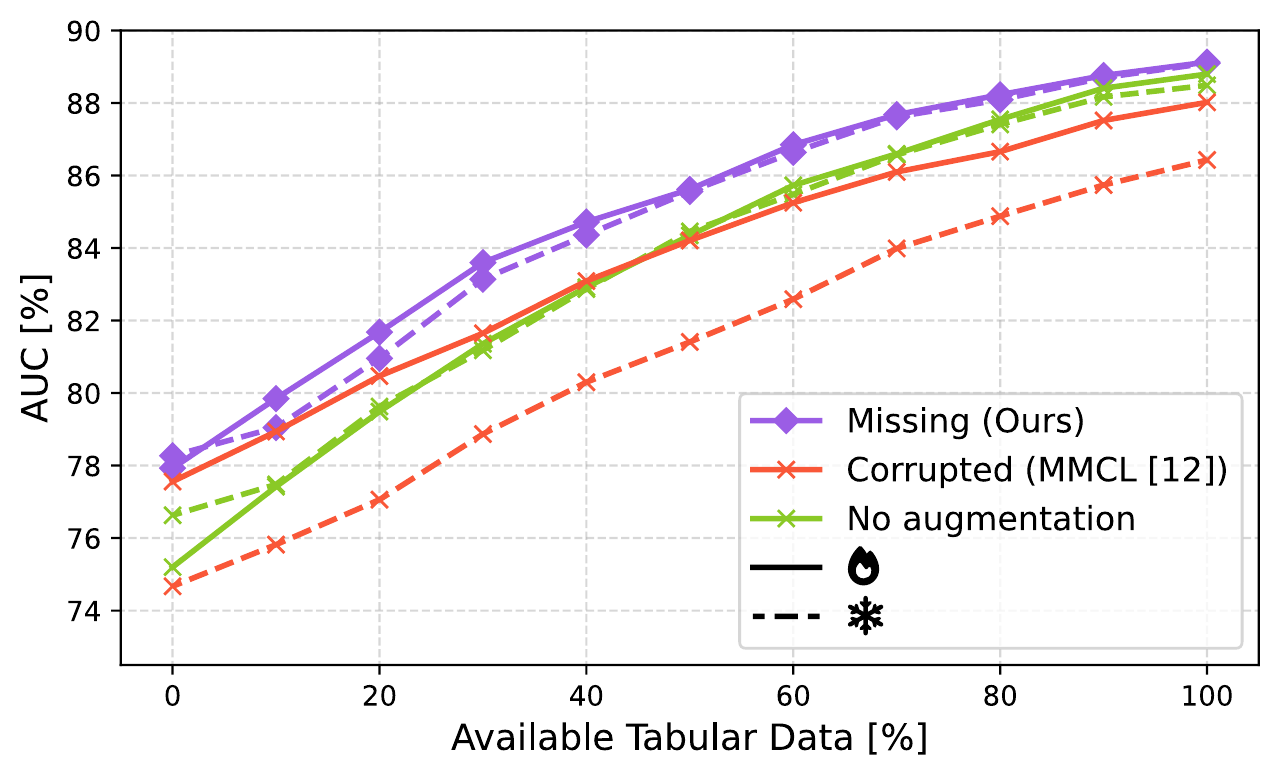}
    \caption{The effect of different tabular augmentation techniques used at pretraining on the result of CAD classification under random value missingness. \faSnowflake: frozen, \faFire: trainable backbones.}
    \label{fig:pretrain_plot}
\end{figure}

To prepare the tabular encoder for handling missing tabular entries, we propose simulating tabular data missingness as an augmentation technique for the pretraining. We compare the proposed missingness augmentation to value corruptions using Gaussian perturbations, sampling from feature's marginal distribution, which has been applied in vision-tabular learning before~\cite{hager2023best}, and no augmentations. The results are presented in Table~\ref{tab_pretrain}. TARTE~\cite{kim2025table} is used as a tabular encoder for all the techniques. We evaluate the performance of the acquired backbones on CAD classification with both image and tabular data and for left ventricular ejection fraction (LVEF) regression using images only. LVEF, being image-derived, does not require tabular data for prediction and thus serves as an ideal target for our unimodal experiment. Our augmentation approach is the only one that consistently improves the no augmentation baseline across all the tasks. Notably, in a frozen setting all the value corruption-based methods fail to reach performance of the no augmentation setting. We attribute this behavior to the misalignment induced by corrupting the tabular attributes. While attribute corruption has been successful in unimodal tabular contrastive learning~\cite{bahri2021scarf}, its application in vision-tabular learning can result in non-factual pairs, where the clinical information in the table no longer represents the ground truth of the patient. This is particularly harmful when applied to critical attributes like sex, smoking status, or disease labels, which fundamentally shift a patient's profile and degrade the learned representations. In contrast, our missingness augmentation provides partial but truthful information, ensuring that cross-modal alignment remains anchored to authentic data, resulting in better performance. Furthermore, as illustrated in Fig. \ref{fig:pretrain_plot}, our augmentation is especially beneficial at low tabular availability, confirming its role in promoting robustness to missing data.
\subsection{Transferability to Attribute-Constrained Data}

\begin{table}
\centering
\scriptsize
    \caption{The results on the multiclass cardiac disease classification on an external cardiac dataset called M\&M's~\cite{campello2021multi}. The results are reported using AUC. \faSnowflake: frozen, \faFire: trainable backbones. }
    \begin{tabular}{lcc}
\toprule
              & \multicolumn{2}{c}{Disease Classification $\uparrow$} \\ \midrule
              & \tiny{\faSnowflake}           & \tiny{\faFire}       \\ \midrule
Interact Fuse \cite{duanmu2020prediction} &         -             &          62.63           \\
DAFT  \cite{wolf2022daft}        &          -            &       57.45              \\
MMCL \cite{hager2023best}         &          64.93            &         65.37            \\
TGV \cite{hasny2025tgv}         &                   51.59   &       63.01             \\
RoVTL (w/o pretraining)       &          -            &    66.50                 \\
\rowcolor{lightgray} RoVTL (Ours)          &   \textbf{76.43}                   &   \textbf{77.56}    \\ \bottomrule             
\end{tabular}

\label{tab_generalize}
\end{table}
% \textbf{Takeaway:} RobTIL, pretrained on UK Biobank, generalizes to an external cardiac dataset, showing promise for image-tabular foundation models.   

Development of tabular foundation models such as TARTE~\cite{kim2025table} can pave the way for multimodal vision-tabular models. We explore the potential of this idea by evaluating our vision and tabular encoders pretrained on the UK Biobank~\cite{sudlow2015uk} dataset on an external cardiac dataset, M\&M's~\cite{campello2021multi}, for the task of cardiac disease classification. This dataset reflects the challenge of the discrepancy in the number of available tabular attributes in the clinics compared to Biobank, with M\&M's consisting of 16 attributes, compared to the 129 used for the pretraining. Due to the rigid design of the tabular encoder in TIP~\cite{du2024tip} we are unable to use TIP's pretrained encoders for the comparison. Thus, we evaluate RoVTL against image-only inference methods such as MMCL~\cite{hager2023best} and TGV~\cite{hasny2025tgv}, end-to-end supervised methods such as IF~\cite{duanmu2020prediction} and DAFT~\cite{wolf2022daft}, and a randomly initialized variant of our model.
% The results are reported as mean AUC over the varying tabular data missingness scenarios. 
The results are presented in Table \ref{tab_generalize}. RoVTL achieves the best performance, outperforming the randomly initialized model by 9.93 with frozen backbones and 11.06 with trainable backbone. These results highlight RoVTL’s potential to use the extensive knowledge from the UK Biobank and successfully transfer it into the clinical setting, where collecting extensive sets of tabular attributes is often impractical.
% generalization ability and underscore its potential as a foundation for developing large-scale medical tabular-vision models.
 % 

\subsection{Ablation Study}

\begin{table}[t]
\centering
\scriptsize
    \caption{Ablation study showing the effectiveness of each part of our method. The element listed as \textit{w/o} is the one being removed. CAD is evaluated using AUC, while BMI using MAE. \faSnowflake: frozen backbones, \faFire: trainable backbones.}
    \begin{tabular}{lcccc}
\toprule
                    & \multicolumn{2}{c}{CAD $\uparrow$} & \multicolumn{2}{c}{BMI $\downarrow$} \\ \midrule
                    & \tiny{\faSnowflake}     & \tiny{\faFire}     & \tiny{\faSnowflake}      & \tiny{\faFire}     \\ \midrule
w/o pretraining     & -           &       82.94    & -           &  1.71         \\
w/o gate            &      84.60       &     82.13      &         1.61    &    1.87       \\
w/o downstream missingness &      77.07       &     83.37      &           4.64  &     4.53      \\
w/o TabMoFe         &     84.19        &     83.09      &          1.54   &     1.98      \\
% w/o DGL             &     84.40        &    84.47       &          1.87   &   1.98        \\
\rowcolor{lightgray} RoVTL (Ours )               &      \textbf{84.68}       &      \textbf{ 84.91}    &        \textbf{1.51}     &     \textbf{1.69}     \\ \bottomrule
\end{tabular}

\label{tab:ablation}
\end{table}

Each element of our method was designed to ensure robustness to missing tabular data entries. To assess the contribution of each component, we conduct an ablation study in which one component is removed at a time. The results are summarized in Table~\ref{tab:ablation}. We use multilabel CAD classification and BMI regression for the ablations. RoVTL with all its components achieves the best results for all the tasks under both frozen and trainable backbones, clearly showing the effectiveness of our method. In addition, our proposed attribute missingness augmentation has the biggest impact in performance, demonstrating the importance of training under diverse tabular data availability scenarios.
\section{Conclusion}

We introduce RoVTL: Robust Vision-Tabular Learning, the first vision-tabular method, to our knowledge, that performs effectively across the full spectrum of tabular data completeness. RoVTL consists of two main components, contrastive pretraining and downstream task tuning. At pretraining, we propose tabular data missingness as an augmentation strategy, simulating incomplete entries to enhance the robustness of the tabular encoder. 
During downstream task tuning, data missingness is further enhanced with the proposed TabFoMe loss, which ranks performance based on available tabular attributes, and a gated cross-attention module that adaptively adjusts tabular contributions.
Our extensive experiments demonstrate that RoVTL outperforms prior methods, with the largest improvement under limited tabular data. Successful evaluation on external medical dataset highlights its potential for vision-tabular foundation models, while results on the 2D car dataset demonstrate its generalizability to multiple domains. Overall, RoVTL provides a generalizable and robust solution for vision-tabular learning, bridging the gap between image-only and multimodal inference and demonstrating effectiveness across the full spectrum of tabular data scenarios.

\section*{Acknowledgments}
This research has been conducted using the UK Biobank Resource under Application Number 87065. MH is in part supported by the Munich School of Data Science (MUDS) and the European Laboratory for Learning and Intelligent Systems (ELLIS) PhD program. LD
and MH are supported by the German Federal Ministry of Research, Technology and Space (DECIPHER-M, 01KD2420G).
\section*{Impact Statement}
We introduce RoVTL, a method for robust vision-tabular learning across the entire spectrum of tabular data availability. The focus of our work is to allow biobank-scale learning, involving hundreds of tabular attributes, to be applicable in clinical practice, where such scale of tabular data is challenging to acquire. Our method can have a broad impact on decision making in clinical practice, allowing image-only models to be aided by tabular data, if such is available, without risking degraded performance. However, it is important to note that our pretrained encoders are trained using the UK Biobank cohort, which represents predominately white population. As a result, downstream use of these models may reflect or amplify demographic biases present in the training data, including potential racial bias. Careful evaluation across diverse populations is therefore necessary before deployment, and additional training or calibration using more representative cohorts may be required to ensure equitable performance.

Further, while our method is evaluated using different notions of missingness, namely random, least important, and most important, it is unclear how the performance would be affected for tasks where attributes might be Missing Not At Random (MNAR). In clinical practice, depending on a task or given hospital, there could be an underlying reason for the attribute to be missing. While we don't find any MNAR attributes for our tasks, such evaluation is necessary before deploying the methods for tasks involving MNAR attributes.  

Overall, RoVTL provides a path towards more robust vision-tabular learning, providing a foundation for developing systems that can adapt to real-world clinical constraints. We plan to extend this method to tasks that involve MNAR scenarios and further validate our models on varying demographics. 
\nocite{bai2018automated}

\bibliography{main}
\bibliographystyle{icml2026}

%%%%%%%%%%%%%%%%%%%%%%%%%%%%%%%%%%%%%%%%%%%%%%%%%%%%%%%%%%%%%%%%%%%%%%%%%%%%%%%
%%%%%%%%%%%%%%%%%%%%%%%%%%%%%%%%%%%%%%%%%%%%%%%%%%%%%%%%%%%%%%%%%%%%%%%%%%%%%%%
% APPENDIX
%%%%%%%%%%%%%%%%%%%%%%%%%%%%%%%%%%%%%%%%%%%%%%%%%%%%%%%%%%%%%%%%%%%%%%%%%%%%%%%
%%%%%%%%%%%%%%%%%%%%%%%%%%%%%%%%%%%%%%%%%%%%%%%%%%%%%%%%%%%%%%%%%%%%%%%%%%%%%%%
\newpage
\appendix
% \onecolumn
\clearpage
\setcounter{page}{1}

\section{Detailed Data Description}

\subsection{Tabular Attributes}

Selection of the tabular attributes in our method is dependent on the training stage. We present comprehensive overview of the attributes, at which stage they are used, and their missingness rates in Table~\ref{supp_tab:ukbb_attr} for UK Biobank~\cite{sudlow2015uk}, Table~\ref{supp_tab:mnms_attr} for M\&Ms~\cite{campello2021multi}, and Table~\ref{supp_tab:dvm_attr} for DVM~\cite{huang2023dvmcarlargescaleautomotivedataset}. As the M\&Ms dataset does not include missing attributes, we do not report the missingness rates. For the UK Biobank we also include the field IDs that are used to extract the attributes. The attributes that are marked as \textit{extracted} are derived using automated cardiac MRI analysis \cite{bai2018automated}.

\begin{table*}[!h]
\centering
\caption{Tabular attributes extracted from the UK Biobank \cite{sudlow2015uk} and their corresponding field IDs. The \faCheck\ symbol in the CAD or BMI column indicates that the attribute is used in the respective downstream task, while the \faTimes\ symbol indicates that it is excluded from that task. All listed attributes are used during pretraining. Miss. describes the percentage (\%) of values originally missing in the given column.}
\label{supp_tab:ukbb_attr}
\tiny
\begin{tabular}{|p{4cm} l c c c  | p{4cm} l c c c|}
\hline
\textbf{Tabular Feature} & \textbf{Field ID} & \textbf{Miss.} & \textbf{CAD} & \textbf{BMI} &
\textbf{Tabular Feature} & \textbf{Field ID} & \textbf{Miss.} & \textbf{CAD} & \textbf{BMI} \\
\hline

Alcohol drinker status & 20117  & 0.7 & \faCheck & \faCheck & Alcohol intake frequency & 1558  & 0.7 & \faCheck & \faCheck \\
Pulse rate automated & 102  & 19.05 & \faCheck & \faCheck & Augmentation index for PWA & 12681 & 10.2 & \faCheck & \faCheck \\
Average heart rate & 22426 & 20.8 & \faCheck & \faCheck & Basal metabolic rate & 23105  & 5.2 & \faCheck & \faCheck \\
Medication for cholesterol, blood pressure or diabetes & 6177  & 7.4 & \faCheck & \faCheck & Body fat percentage & 23099  & 5.2 & \faCheck & \faCheck \\
Body mass index & 21001 & 3.2 & \faCheck & \faTimes & Body surface area & 22427 & 20.9 & \faCheck & \faCheck \\
Cardiac index & 22425 & 20.9 & \faCheck & \faCheck & Cardiac index during PWA & 12702  & 17.4 & \faCheck & \faCheck \\
Cardiac operations performed & 20004  & 26.4 & \faCheck & \faCheck & Cardiac output & 22424  & 20.8 & \faCheck & \faCheck \\
Cardiac output during PWA & 12682  & 13.6 & \faCheck & \faCheck & Central augmentation pressure during PWA & 12680  & 10.8 & \faCheck & \faCheck \\
Central pulse pressure during PWA & 12678  & 10.7 & \faCheck & \faCheck & Central systolic blood pressure during PWA & 12677  & 10.7 & \faCheck & \faCheck \\
Cholesterol lowering medication regularly taken & 6177  & 52.1 & \faCheck & \faCheck & Current tobacco smoking & 1239   & 7.4 & \faCheck & \faCheck \\
Diastolic blood pressure automated & 4079 & 19.0 & \faCheck & \faCheck & Diastolic blood pressure manual reading & 94 & 92.7 & \faCheck & \faCheck \\
Diastolic brachial blood pressure during PWA & 12675  & 10.7 & \faCheck & \faCheck & Duration of heavy DIY & 2634  & 53.7 & \faCheck & \faCheck \\
Duration of moderate activity & 894  & 10.2 & \faCheck & \faCheck & Duration of strenuous sports & 10010  & 87.8 & \faCheck & \faCheck \\
Duration of vigorous activity & 914  & 32.6 & \faCheck & \faCheck & Duration of walks & 874 & 2.3 & \faCheck & \faCheck \\
Duration walking for pleasure & 981  & 19.2 & \faCheck & \faCheck & End systolic pressure during PWA & 12683  & 10.9 & \faCheck & \faCheck \\
End systolic pressure index during PWA & 12684 & 13.6 & \faCheck & \faCheck & Ever smoked & 20160 & 1.1 & \faCheck & \faCheck \\
Exposure to tobacco smoke at home & 1269  & 2.7 & \faCheck & \faCheck & Exposure to tobacco smoke outside home & 1279  & 2.7 & \faCheck & \faCheck \\
Falls in the last year & 2296  & 0.7 & \faCheck & \faCheck & Frequency of other exercises in last 4 weeks & 3637  & 45.6 & \faCheck & \faCheck \\
Heart rate during PWA & 12673  & 10.2 & \faCheck & \faCheck & Height & 12144  & 0.0 & \faCheck & \faTimes \\
Hip circumference & 49  & 2.9 & \faCheck & \faCheck & Hormone replacement therapy medication regularly taken & 6153  & 48.6 & \faCheck & \faCheck \\
Mean arterial pressure during PWA & 12687 & 10.7 & \faCheck & \faCheck & Number of beats in waveform average for PWA & 12679  & 10.2 & \faCheck & \faCheck \\
Days/week moderate activity & 884  & 0.7 & \faCheck & \faCheck & Days/week vigorous activity & 904  & 0.7 & \faCheck & \faCheck \\
Days/week walked & 864  & 0.7 & \faCheck & \faCheck & Overall health rating & 2178  & 0.7 & \faCheck & \faCheck \\
P duration & 12338  & 16.5 & \faCheck & \faCheck & Pace & 3079  & 2.9 & \faCheck & \faCheck \\
Pack years adult smoking as proportion of lifespan exposed & 20162  & 76.5 & \faCheck & \faCheck & Pack years of smoking & 20161  & 76.5 & \faCheck & \faCheck \\
Past tobacco smoking & 1249  & 2.7 & \faCheck & \faCheck & Peripheral pulse pressure during PWA & 12676  & 10.7 &  \faCheck & \faCheck \\
Pulse wave Arterial Stiffness index & 21021  & 16.0 & \faCheck & \faCheck & QRS duration & 12340  & 12.1 & \faCheck & \faCheck \\
Shortness of breath walking on level ground & 4717  & 0.7 & \faCheck & \faCheck & Sleep duration & 1160  & 0.7 & \faCheck & \faCheck \\
Sleeplessness / insomnia & 1200  & 0.7 & \faCheck & \faCheck & Smoking/smokers in household & 1259  & 2.7 & \faCheck & \faCheck \\
Stroke volume during PWA & 12686  & 13.5 & \faCheck & \faCheck & Systolic blood pressure & 4080  & 19.0 & \faCheck & \faCheck \\
Systolic brachial blood pressure during PWA & 12674  & 10.7 & \faCheck & \faCheck & Total mass & 23283  & 24.6 & \faCheck & \faCheck \\
Total peripheral resistance during PWA & 12685  & 10.9 & \faCheck & \faCheck & Ventricular rate & 12336  & 12.1 & \faCheck & \faCheck \\
Waist circumference & 48  & 2.9 & \faCheck & \faCheck & Weight & 21002  & 3.1 & \faCheck & \faTimes \\
Whole body fat-free mass & 23101 & 5.2  & \faCheck & \faCheck & Whole body fat mass & 23100 & 5.3 & \faCheck & \faCheck \\
Whole body water mass & 23102 & 5.2 & \faCheck & \faCheck & Worrier / anxious feelings & 1980 & 0.7 & \faCheck & \faCheck \\
Date of birth (Age) & 33 & 0.0 & \faCheck & \faCheck & Sex & 31 & 0.0 & \faCheck & \faCheck \\
Heart Failure (I50) & 13135 & 0.0 & \faTimes & \faTimes & Atrial Fibrillation and Flutter (I48) & 13135 & 0.0 & \faTimes & \faTimes \\
Nonrheumatic Aortic Valve Disorders (I25) & 131324  & 0.0 & \faTimes & \faTimes & Other Diseases of Pericardium (I31) & 131316  & 0.0 & \faTimes & \faTimes \\
Paroxysmal Tachycardia (I47) & 131348  & 0.0 & \faTimes & \faTimes & Acute Pericarditis (I30) & 131314  & 0.0 & \faTimes & \faTimes \\
Atrioventricular and Left Bundle-Branch Block (I44) & 131342  & 0.0 & \faTimes & \faTimes & Nonrheumatic Mitral Valve Disorders (I34) & 131322  & 0.0 & \faTimes & \faTimes \\
Acute and Subacute Endocarditis (I33) & 131320  & 0.0 & \faTimes & \faTimes & Other Cardiac Arrhythmias (I49) & 131352  & 0.0 & \faTimes & \faTimes \\
Cardiomyopathy (I42) & 131338 & 0.0 & \faTimes & \faTimes & Acute Myocarditis (I40) & 131334 & 0.0 & \faTimes & \faTimes \\
Other Conduction Disorders (I45) & 131344 & 0.0 & \faTimes & \faTimes & Cardiac Arrest (I46) & 131346 & 0.0 & \faTimes & \faTimes \\
Nonrheumatic Tricuspid Valve Disorders (I36) & 131326 & 0.0 & \faTimes & \faTimes & Essential (Primary) Hypertension (I10) & 131286 & 0.0 & \faTimes & \faTimes \\
Hypertensive Heart Disease (I11) & 131288 & 0.0 & \faTimes & \faTimes & Hypertensive Renal Disease (I12) & 131290 & 0.0 & \faTimes & \faTimes \\
Hypertensive Heart and Renal Disease (I13) & 131292 & 0.0 & \faTimes & \faTimes & Secondary Hypertension (I15) & 131294 & 0.0 & \faTimes & \faTimes \\
Insulin-Dependent Diabetes Mellitus (E10) & 130706 & 0.0 & \faTimes & \faTimes & Non-Insulin-Dependent Diabetes Mellitus (E11) & 130708 & 0.0 & \faTimes & \faTimes \\
Other Specified Diabetes Mellitus (E13) & 130712 & 0.0 & \faTimes & \faTimes & Unspecified Diabetes Mellitus (E14) & 130714 & 0.0 & \faTimes & \faTimes \\
Angina Pectoris (I20) & 131296 & 0.0 & \faTimes & \faTimes & Other Acute Ischaemic Heart Diseases (I24) & 131304 & 0.0 & \faTimes & \faTimes \\
Chronic Ischaemic Heart Disease (I25) & 131306  & 0.0 & \faTimes & \faTimes & Myocardial Infarction (I21) & 42000 & 0.0 & \faTimes & \faTimes \\
Left Ventricular Ejection Fraction & \textit{Extracted}  & 0.1 & \faCheck & \faCheck & Left Ventricular End-diastolic Volume & \textit{Extracted}  &  0.0 & \faCheck & \faCheck \\
Left Ventricular End-systolic Volume & \textit{Extracted}  & 0.0 & \faCheck & \faCheck & Left Ventricular Stroke Volume & \textit{Extracted}  & 0.0 & \faCheck & \faCheck \\
Left Ventricular End-diastolic Mass & \textit{Extracted}  & 0.0 & \faCheck & \faCheck & Left Ventricular Cardiac Output & \textit{Extracted}  & 0.0 & \faCheck & \faCheck \\
Right Ventricular End-diastolic Volume & \textit{Extracted}  & 0.0 & \faCheck & \faCheck & Right Ventricular End-systolic Volume & \textit{Extracted}  & 0.0 & \faCheck & \faCheck \\
Right Ventricular Stroke Volume & \textit{Extracted} & 0.0 & \faCheck & \faCheck & Right Ventricular Ejection Fraction & \textit{Extracted} & 0.1 & \faCheck & \faCheck \\
Right Ventricular Cardiac Output & \textit{Extracted} & 0.0 & \faCheck & \faCheck & Myocardial End-systolic Volume & \textit{Extracted} & 0.0 & \faCheck & \faCheck \\
Myocardial End-diastolic Volume & \textit{Extracted} & 0.0 & \faCheck & \faCheck & & & & & \\

\hline
\end{tabular}
\end{table*}

\subsection{ICD Codes}
\label{sup_sec:icd}
The labels for multilabel cardiovascular artery disease (CAD) classification task are defined using the ICD-10 codes of categories I20-I25. Diseases defined by the following codes are used as positive labels:
\begin{itemize}
    \vspace{2pt}
    \item Angina pectoris: I20.0, I20.2, I20.8, I20.9;
    % \vspace{2pt}
    \item  Acute myocardial infarction: I21.0, I21.1, I21.2, I21.3, I21.4, I21.9;
    \item Other acute ischemic heart disease: I24.0, I24.8, I24.9;
    % \vspace{2pt}
    \item Chronic ischemic heart disease: I25.0, I25.1, I25.2, I25.3, I25.4, I25.5, I25.6, I25.8, I25.9.
    \vspace{2pt}
\end{itemize}
We extract the date of each MR scan and consider the subject CAD positive only if the disease was reported prior to the scanning date. UK Biobank uses self-reporting of diseases and due to the nature of CAD, which can remain undiagnosed for years, it is possible that some of the healthy subjects do have CAD. Addressing the label uncertainty in UK Biobank is left for future work. 

\subsection{M\&M's Cardiac Disease Classification}
The cardiac diseases included in the M\&M's dataset \cite{campello2021multi} differ from those used for the CAD classification in UK Biobank \cite{sudlow2015uk}. M\&M's includes patients representing four cardiac diseases and healthy samples. The diseases are:
\begin{itemize}
    \vspace{2pt}
    \item Hypertrophic cardiomyopathy (HCM),
    \vspace{2pt}
    \item Dilated cardiomyopathy (DCM),
    \vspace{2pt}
    \item Hypertensive Heart Disease (HHD),
    \vspace{2pt}
    \item Abnormal Right Ventricle (ARV).
    \vspace{2pt}
\end{itemize}

\section{Implementation Details}
\label{sup_sec:details}

\subsection{Image Augmentations}
Image augmentations are applied using the torchvision library. Due to different data formats (2D vs. 3D) between DVM \cite{huang2023dvmcarlargescaleautomotivedataset} and cardiac datasets, UK Biobank \cite{sudlow2015uk} and M\&M's \cite{campello2021multi}, different augmentations are applied to them. For UK biobank and M\&M's the following augmentations are applied:
\begin{itemize}
\vspace{2pt}
\item RandomHorizontalFlip(probability=0.5),
\vspace{2pt}
\item RandomResizedCrop(size=128, scale=(0.6, 1.0)),
\vspace{2pt}
\item RandomRotation(degrees=45).
\vspace{2pt}
\end{itemize}
For DVM the following image augmentations are used:
\begin{itemize}
    \vspace{2pt}
    \item ColorJitter(brightness=0.8, contrast=0.8, saturation=0.8, probability=0.8),
    \vspace{2pt}
    \item RandomGrayscale(probability=0.2),
    \vspace{2pt}
    \item GaussianBlur(kernel\_size=29, sigma=(0.1, 2.0), probability=0.5),
    \vspace{2pt}
    \item RandomResizedCrop(size=128, scale=(0.6, 1.0), ratio=(0.75, 1.33)),
    \vspace{2pt}
    \item RandomHorizontalFlip(probability=0.5),
    \vspace{2pt}
\end{itemize}

\subsection{Tabular Data Augmentations}

% \noindent \textbf{Tabular Data Corruption. } In our pretraining experiments assessing the impact of different augmentation strategies, we adopt the tabular data corruption scheme as described in MMCL \cite{hager2023best}, using an augmentation rate of 0.3. In this scheme, randomly selected entries are replaced with alternative values drawn from the empirical marginal distribution of their respective columns, that is, values observed elsewhere in the dataset. 

\noindent \textbf{Tabular Data Corruption.} In our pretraining experiments assessing the impact of different augmentation strategies, we adopt two corruption schemes with an augmentation rate of 0.3, follwoing prior works~\cite{hager2023best}. The first corruption follows the marginal distribution sampling described in \cite{hager2023best}, where randomly selected entries are replaced with values observed elsewhere in the dataset. The second is a hybrid Gaussian corruption strategy: for a subset of features, numerical values are perturbed with additive Gaussian noise $\mathcal{N}(0, 0.5^2)$, while categorical features are subjected to random replacement with a different category observed in the column. In both schemes, the original value is replaced or modified to simulate data noise, and any pre-existing missing values are preserved to maintain the original sparsity pattern.

\subsection{Tuning with Disentangled Gradient Learning} \label{supp:dgl}
A common challenge in multimodal learning is under-optimization, where joint training can lead to lower performance compared to unimodal models. One of the techniques addressing this issue is disentangled gradient learning DGL~\cite{wei2025boosting}. We introduce this training strategy into our downstream task learning to ensure effective optimization of the unimodal models and the multimodal fusion module.

DGL explicitly decouples the optimization of modality encoders from that of the multimodal fusion module using two backward passes. In the first pass, a unimodal loss is computed for each modality, where separate task heads are attached to modality encoders,
\begin{equation}
    \mathcal{L}_{unimodal}=\mathcal{L}_{task}(i) + \mathcal{L}_{task}(t)
\end{equation}
This unimodal loss is used to update each modality encoder independently. The gradients that accumulate on the multimodal fusion parameters during this step are removed, and the modality features are detached so that the upcoming multimodal loss does not propagate through them. In the second pass, the multimodal loss \(\mathcal{L}_{multi}\) is computed and backpropagated only through the multimodal fusion module. This disentanglement resolves the gradient interference between unimodal and multimodal objectives, ensuring stable and balanced optimization across modalities.

\subsection{Baselines} \label{supp:baselines}

For the baselines, we follow the settings and specifications provided by the original authors. Any details specific to our implementations are described below.

\vspace{2pt}
\noindent \textbf{Supervised baselines. } The end-to-end supervised baselines, which include Interact Fuse (IF) \cite{duanmu2020prediction} and DAFT \cite{wolf2022daft} are trained using the attributes as listed for each task in Tables~\ref{supp_tab:ukbb_attr}-~\ref{supp_tab:dvm_attr}, without any pretraining including the remaining attributes. This approach ensures no leakage of task specific information, as pretraining incorporated attributes that are later used as labels for downstream tasks. The missing values are handled by concatenating the input attributes with a mask describing which attributes are present and which ones are missing. Zero imputation is used for the missing values. 

\vspace{2pt}
\noindent \textbf{TIP \cite{du2024tip}. } The original TIP pipeline does not include pretraining with labels, which was introduced in MMCL \cite{hager2023best}. To ensure proper comparison to other pretraining based approaches we use the same set of attributes, which also incorporate labels. For TIP, we use fixed masks at downstream task tuning that mask the attributes that should be excluded for the given downstream task prediction. As the TIP tabular encoder expects the attributes at certain order, we could not simply exclude the values and pad the remaining tokens, motivating the use of masking which was already implemented in the method as a way of marking missing entries. For the edge-case of tabular data being fully missing, we follow the typical TIP prediction pipeline, but mask all the tabular tokens. 

\vspace{2pt}
\noindent \textbf{Concat Fuse (CF) \cite{spasov2019parameter} and Max Fuse (MF) \cite{vale2021long}. } We use two different multimodal embedding concatenation methods as baselines. Namely, CF where tabular and image embeddings are concatenated together, and MF, where element-wise max pooling is performed across two embeddings, selecting the larger values in each feature dimension to form the multimodal embedding. To ensure proper comparison, we use our pretrained image and tabular encoders for the two concatenation approaches. Furthermore, we compare RoVTL against those methods tuned using our downstream task tuning approach and the standard setting, without incorporating TabMoFe loss, missing value simulation, and DGL \cite{wei2025boosting}. 
% performs element-wise max pooling across the two embeddings, selecting the larger value in each feature dimension to form the multimodal representation.
\begin{table}[t]
\centering
\caption{Full list of the tabular attributes from the M\&M's dataset \cite{campello2021multi}. All attributes listed are used for the downstream task (cardiac disease classification), as denoted by \faCheck\ .}
\label{supp_tab:mnms_attr}
\tiny
\begin{tabular}{|lc|}
\hline
\textbf{Tabular Feature} & \textbf{Classification} \\ \hline
Left Ventricular End-Diastolic Volume  & \faCheck \\ 
Left Ventricular End-Systolic Volume  & \faCheck \\ 
Left Ventricular Stroke Volume  & \faCheck \\ 
Left Ventricular Ejection Fraction  & \faCheck \\ 
Left Ventricular End-Diastolic Mass  & \faCheck \\ 
Left Ventricular Cardiac Output  & \faCheck \\

Right Ventricular End-Diastolic Volume  & \faCheck \\ 
Right Ventricular End-Systolic Volume  & \faCheck \\ 
Right Ventricular Stroke Volume  & \faCheck \\ 
Right Ventricular Ejection Fraction  & \faCheck \\
Right Ventricular Cardiac Output  & \faCheck \\ 

Myocardial End-Diastolic Volume & \faCheck \\ 
Myocardial End-Systolic Volume  & \faCheck \\ 

Height & \faCheck \\ 
Weight & \faCheck \\
Sex & \faCheck \\ \hline

\end{tabular}
\end{table}

\begin{table}[t]
\centering
\caption{A list of tabular attributes from the DVM dataset \cite{huang2023dvmcarlargescaleautomotivedataset}. All the attributes are used for the pretraining. The attributes denoted by \faCheck\ in the \textit{Classification} column are used in the downstream task prediction (car model classification), while the ones marked by \faTimes\ are excluded. Miss. describes the percentage (\%) of values originally missing in the given column.}
\label{supp_tab:dvm_attr}
\tiny
\begin{tabular}{|ccc|}
\hline
\textbf{Tabular Feature} & \textbf{Miss. (\%)} & \textbf{Classification} \\ \hline
    Manufacturer Name         & 0.0  &      \faTimes          \\ 
      Car Model          & 0.0 &      \faTimes          \\ 
      Color          & 8.1 &    \faCheck            \\ 
     Registration Year        & 0.0  &         \faCheck       \\ 
      Body Type     &   3.6  &         \faCheck       \\ 
      Mileage Miles     & 0.4    &        \faCheck        \\ 
    Engine Size    & 0.8  &  \faCheck              \\ 
      Gearbox      & 0.1  &        \faCheck        \\ 
      Fuel Type     & 0.2   &        \faCheck        \\ 
      Price       & 0.4 &        \faCheck        \\ 
      Engine Power   &  11.86    &        \faTimes      \\ 
      Annual Tax   &   17.4  &        \faTimes     \\ 
      Wheelbase    &   1.7  &        \faTimes       \\ 
      Length      &  0.3 &        \faTimes       \\ 
      Height     &  0.4  &        \faTimes       \\ 
      Width     &  0.3  &        \faTimes       \\ 
      Average Miles per Gallon   &  15.63   &        \faTimes       \\ 
      Top Speed   &   16.25   &        \faTimes       \\ 
      Seat Count   &  2.4    &        \faCheck        \\ 
      Door Count    &   1.7  &        \faCheck        \\ 
          \hline
\end{tabular}
\end{table}

% {
%     "Maker": 0.0,
%     "Genmodel": 0.0,
%     "Genmodel_ID": 0.0,
%     "Adv_ID": 0.0,
%     "Adv_year": 0.0,
%     "Adv_month": 0.0,
%     "Color": 0.08154554435145664,
%     "Reg_year": 2.6094574192466124e-05,
%     "Bodytype": 0.003556317682801812,
%     "Runned_Miles": 0.003932825110435966,
%     "Engin_size": 0.007694171590464297,
%     "Gearbox": 0.0006225419843059775,
%     "Fuel_type": 0.0015246686921026634,
%     "Price": 0.004268326778624816,
%     "Engine_power": 0.1185849285195057,
%     "Annual_Tax": 0.1739948929190509,
%     "Wheelbase": 0.0169689288177294,
%     "Height": 0.0037128851279566084,
%     "Width": 0.0032133604219865427,
%     "Length": 0.003157443477288401,
%     "Average_mpg": 0.1563176828018117,
%     "Top_speed": 0.16253937484855827,
%     "Seat_num": 0.024133753331717956,
%     "Door_num": 0.016972656614042608
% }

\subsection{Downstream Task Tuning}

We perform downstream task tuning under two settings, with frozen and trainable backbones. The multimodal fusion module is trainable in every setting and is added on top of the pretrained backbones. The Adam optimizer with a weight decay of \(1e^{-4}\) and an effective batch size of 512 for DVM \cite{huang2023dvmcarlargescaleautomotivedataset} and 256 for UK Biobank \cite{sudlow2015uk} is used. We perform a learning rate sweep covering four values, \(\{1e^{-3}, 3e^{-3}, 1e^{-4}, \text{and } 3e^{-4}\}\), with the best one selected based on the validation set results. The tuning is performed for 500 epochs for the DVM dataset, 100 for M\&Ms dataset \cite{campello2021multi}, and 100 for UK Biobank. We use the Huber loss for regression (BMI regression), Binary Cross Entropy with Logits loss for multilabel classification (DVM car model prediction and M\&Ms cardiac disease prediction), and Cross Entropy loss for multiclass classification (CAD classification on UK Biobank). The augmentation rate is set to 0.95. 

\section{Additional Experiments}

\subsection{Robustness to Missing Entries (Complete)}
\label{sup_sec:missing_entries}

\begin{figure*}[]
    \centering
    \includegraphics[width=\textwidth]{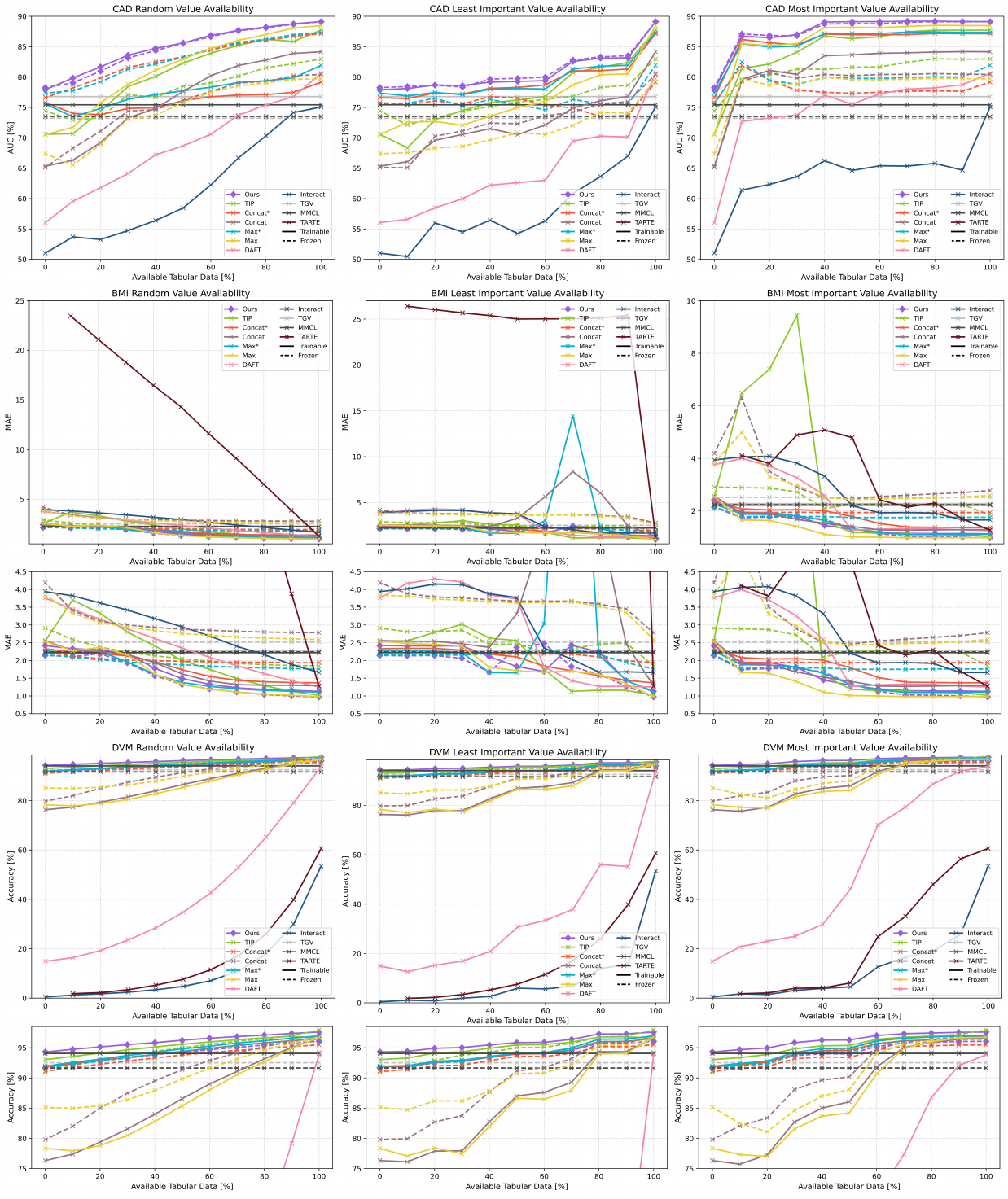}
    \caption{Model performance plotted over tabular data availability from 0\% to 100\%. CAD is reported using AUC ($\uparrow$), BMI using MAE ($\downarrow$), and DVM car classification using accuracy ($\uparrow$). For clarity, zoomed-in versions are provided for BMI regression and car model classification with DVM to highlight the performance differences between the models. \faSnowflake: frozen backbones, \faFire: trainable backbones.}
    \label{fig:complete_robust}
\end{figure*}

We provide full results for robustness to missing entries in Fig.~\ref{fig:complete_robust}. For clarity, we omitted baselines with the lowest performance in the main body of the paper. In addition to the baselines, we also include TARTE \cite{kim2025table}, the tabular encoder that we use for RoVTL, as a tabular-only model for comparison. As TARTE does not support native support for multilabel classification, we only include it as a baseline for BMI regression and car model classification. RoVTL outperforms the other methods on all tasks. Furthermore, incorporating our downstream fine-tuning setting to CF and MF improves the robustness of both of the methods to limited tabular data regimes, further showing the strength of our introduced training pipeline. RoVTL consistently outperforms TARTE, demonstrating greater robustness to missing data and underscoring the benefits of combining vision and tabular information.

\begin{figure}[t]
    \centering
    \includegraphics[width=\linewidth]{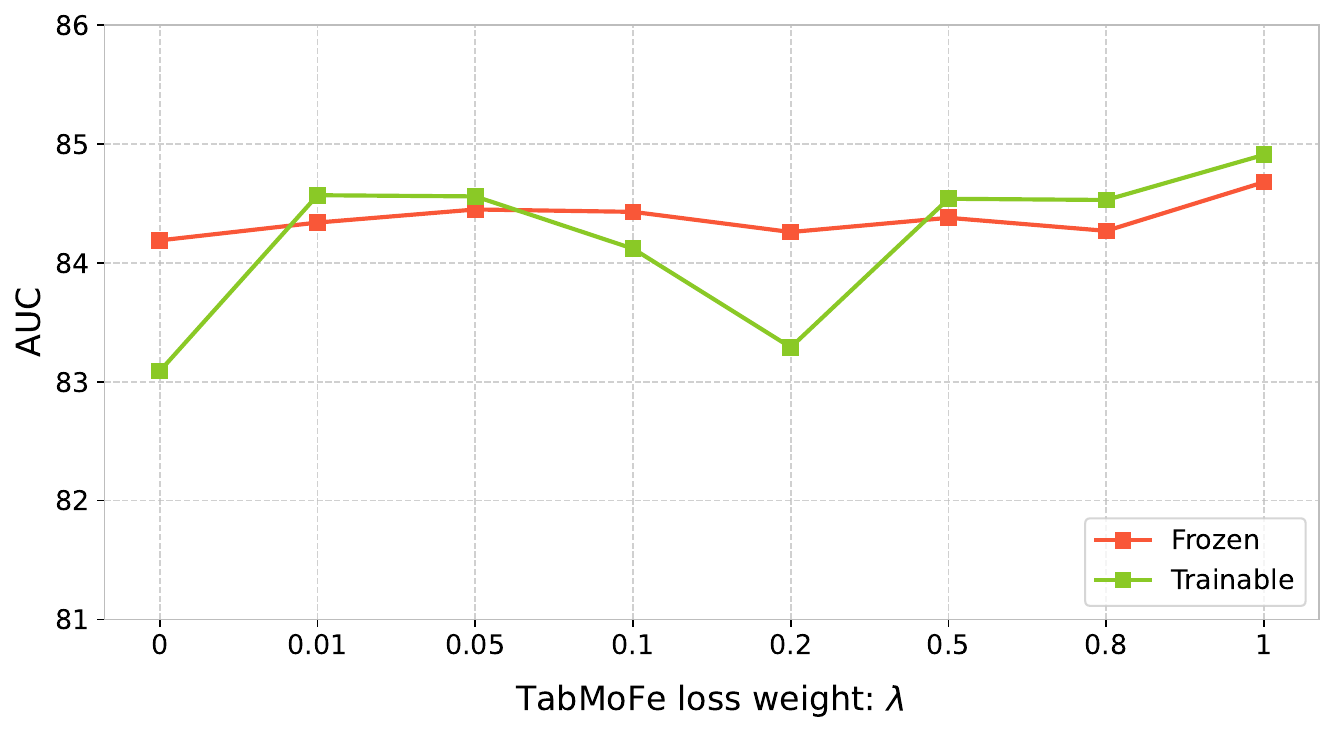}
    \caption{Impact of the TabMoFe loss weighting parameter $\lambda$ on the CAD classification performance (AUC $\uparrow$).}
    \label{fig:hyperparam}
\end{figure}

\subsection{$\lambda$-hyperparameter Analysis} \label{supp:hyperparam}

We conduct an analysis of the effect of the TabMoFe loss weighting parameter $\lambda$ on CAD classification performance. The results, illustrated in Fig.~\ref{fig:hyperparam}, show that aside from the $\lambda=0$ baseline, where the TabMoFe loss is disabled, the model exhibits minimal performance variance across a wide range of values. These results demonstrate that our method is not sensitive to the specific selection of the $\lambda$ hyperparameter, while confirming that the inclusion of the loss function consistently yields a positive effect on overall performance.

\subsection{Interpretability}
\label{sup_sec:interpret}
\begin{figure*}[]
    \centering
    \rotatebox{270}{%
        \includegraphics[width=0.95\textheight]{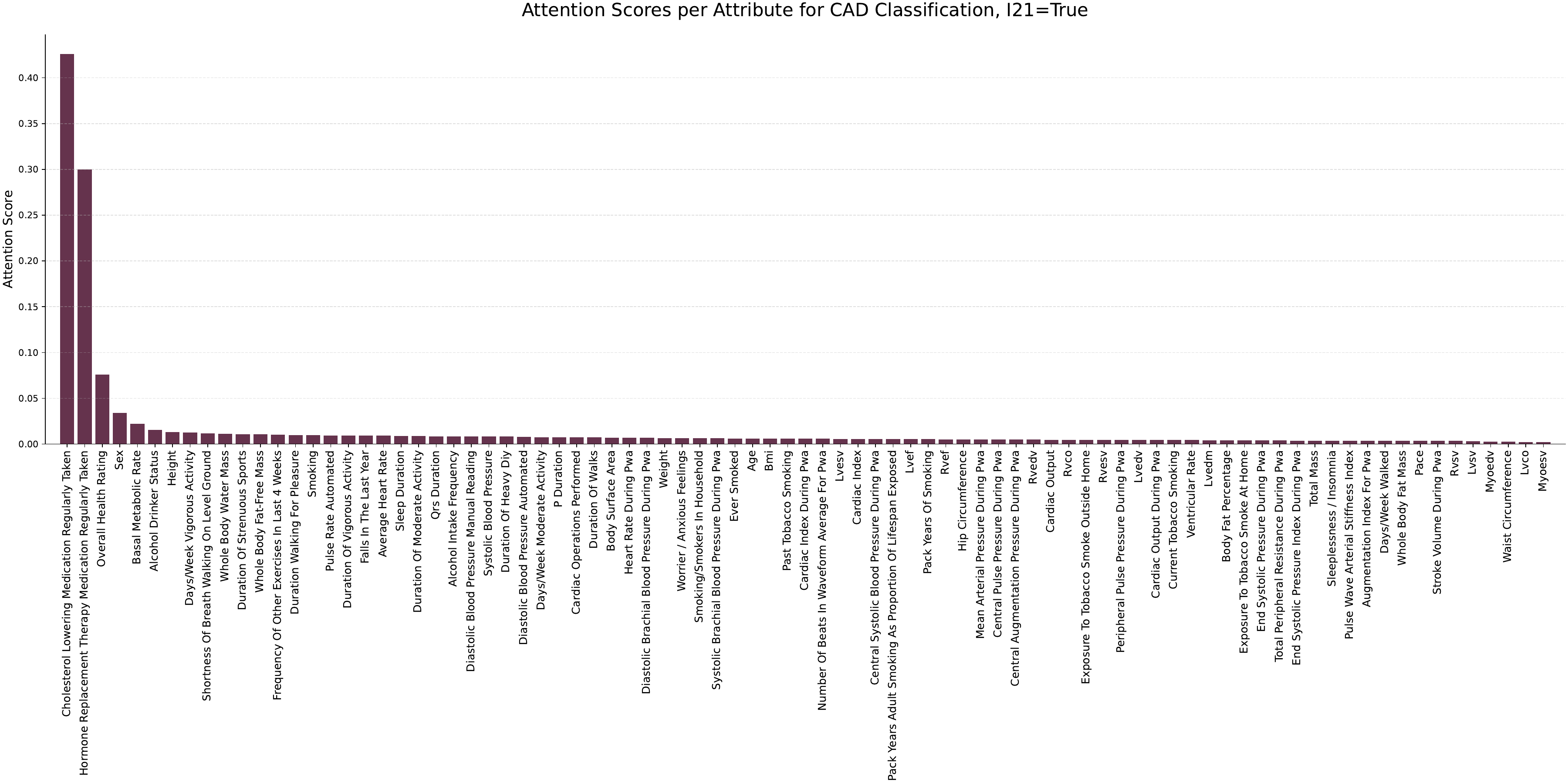}
    }
    \caption{Attention scores per attribute for CAD classification with the UK Biobank. }
    \label{fig:attn_cad}
\end{figure*}

\begin{figure*}[]
    \centering
    \rotatebox{270}{%
        \includegraphics[width=0.95\textheight]{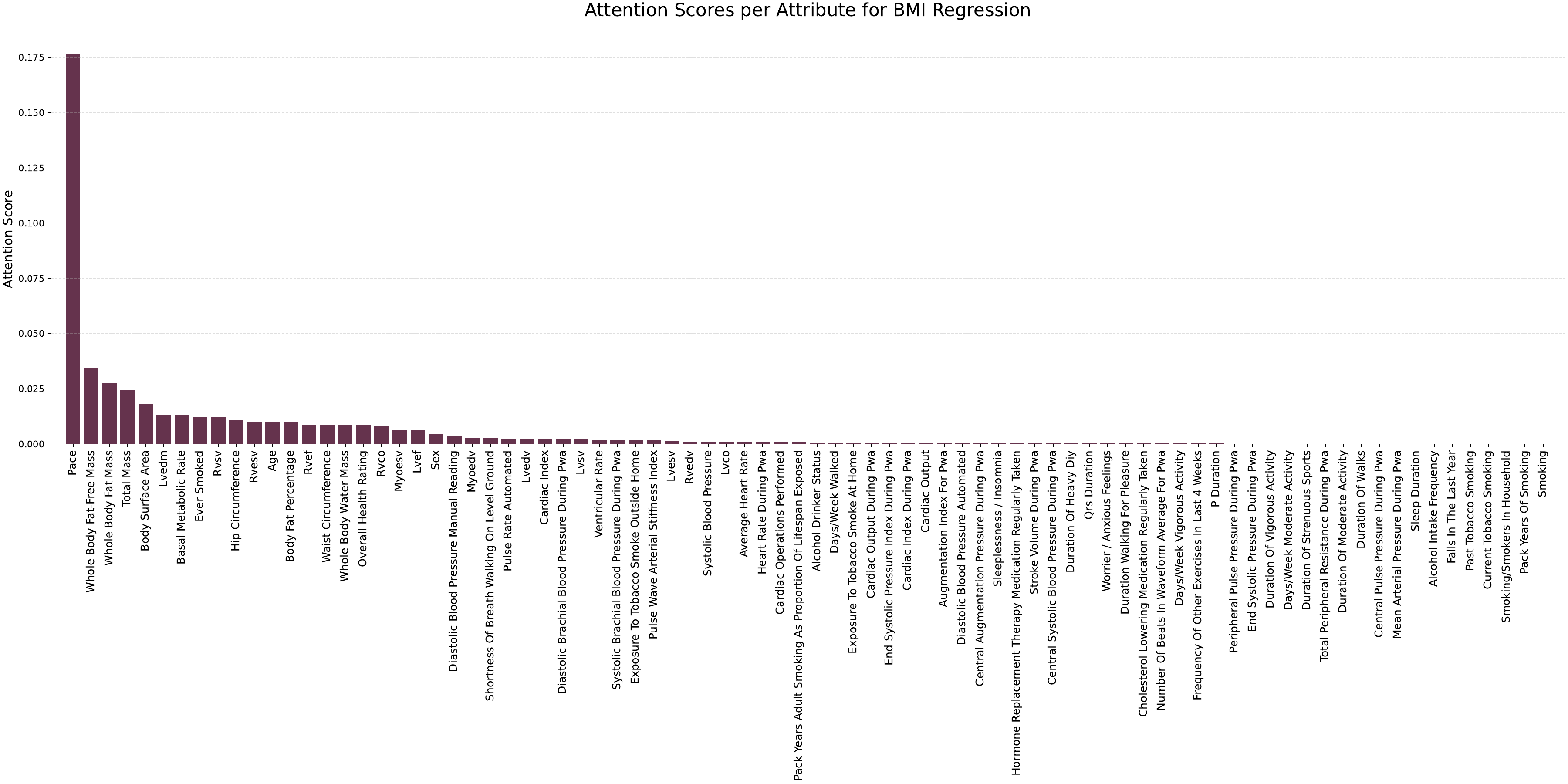}
    }
    \caption{Attention scores per attribute for BMI regression with the UK Biobank. }
    \label{fig:attn_bmi}
\end{figure*}

\begin{figure*}[]
    \centering
    \rotatebox{270}{%
        \includegraphics[width=0.5\textheight]{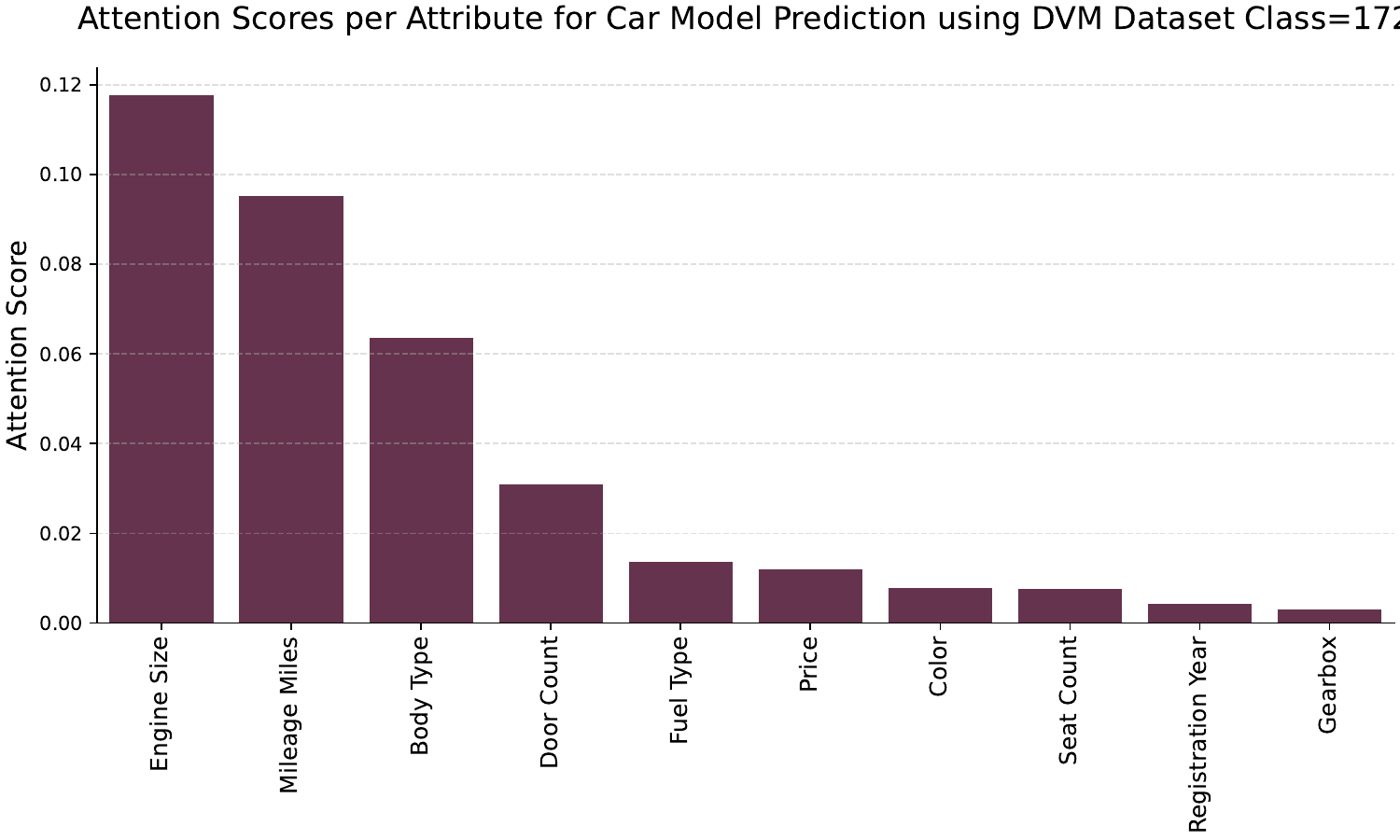}
    }
    \caption{Attention scores per attribute for BMI regression with the UK Biobank. }
    \label{fig:attn_dvm}
\end{figure*}

We evaluate the tabular attributes attention scores to better understand which features RoVTL uses to make predictions. The results are shown in Fig.~\ref{fig:attn_cad} for CAD classification, Fig.~\ref{fig:attn_bmi} for BMI regression, and in Fig.~\ref{fig:attn_dvm} for car model prediction. To obtain the attention scores, the raw attention scores are extracted from the final cross-attention layer. The attention weights assigned to each attribute are then aggregated by first taking the mean over across all heads and then over the image tokens, which produces a single score for each tabular attribute. The aggregation is performed per class for classification and on full test set for regression. RoVTL takes advantage of the complementary information in the tabular attributes that is not directly visible on the image, such as the engine size or mileage of the car, to make predictions. For BMI regression, attributes describing the body, such as the whole body fat mass or whole body fat-free mass, are attended to the most. This behavior confirms that the model does not treat the tabular data as auxiliary, but instead learns to rely on tabular features that provide predictive value beyond what is visible in the images, resulting in more robust and interpretable multimodal predictions.

\clearpage

\end{document}